\documentclass{article}
\usepackage[utf8]{inputenc}
\usepackage{hyperref}
\usepackage{float}
\usepackage{amsmath}
\usepackage{graphicx}
\usepackage{multirow}
\usepackage{caption}
\usepackage{subcaption}
\usepackage{algpseudocode}
\usepackage[ruled,vlined, linesnumbered]{algorithm2e}
\usepackage{booktabs}
\usepackage{longtable}
\usepackage{apacite}
\usepackage{natbib}
\usepackage{breakcites}
\usepackage[dvipsnames]{xcolor}
\usepackage{authblk}
\usepackage[dvipsnames]{xcolor}
\usepackage[export]{adjustbox}

\title{Bagging Supervised Autoencoder Classifier for Credit Scoring}

\begin{document}
\author[1]{Mahsan Abdoli \thanks{\texttt{m.abdoli@aut.ac.ir} (Mahsan Abdoli)}}
\author[2]{Mohammad Akbari \thanks{Corresponding author: \texttt{akbari.ma@aut.ac.ir} (Mohammad Akbari)}}
\author[1]{Jamal Shahrabi \thanks{\texttt{jamalshahrabi@aut.ac.ir} (Jamal Shahrabi)}}
\date{}
\affil[1]{Department of Industrial Engineering, Amirkabir University of Technology (Tehran Polytechnic)}
\affil[2]{Department of Mathematics and Computer Science, Amirkabir University of Technology (Tehran Polytechnic)\vspace{-2.5em}}
\maketitle
\begin{abstract}
Credit scoring models, which are among the most potent risk management tools that banks and financial institutes rely on, have been a popular subject for research in the past few decades. Accordingly, many approaches have been developed to address the challenges in classifying loan applicants and improve and facilitate decision-making. The imbalanced nature of credit scoring datasets, as well as the heterogeneous nature of features in credit scoring datasets pose difficulties in developing and implementing effective credit scoring models, targeting the generalization power of classification models on unseen data. In this paper, we propose the Bagging Supervised Autoencoder Classifier (BSAC) that mainly leverages the superior performance of the Supervised Autoencoder, which learns low-dimensional embeddings of the input data exclusively with regards to the ultimate classification task of credit scoring, based on the principles of multi-task learning. BSAC also addresses the data imbalance problem by employing a variant of the Bagging process based on the undersampling of the majority class. The obtained results from our experiments on the benchmark and real-life credit scoring datasets illustrate the robustness and effectiveness of the Bagging Supervised Autoencoder Classifier in the classification of loan applicants that can be regarded as a positive development in credit scoring models.

\end{abstract}
\section{Introduction}\label{intro}
Credit scoring is one of the core elements of the financial industry that attempts to assess and measure the creditworthiness of loan applicants and classify them into reliable and unreliable ones, aiming to facilitate the decision-making process of banks and financial institutes on granting loans to customers.  In general, robust and trustworthy credit scoring models lead to reduction of administrative costs, minimization of human error or bias in accepting or refusing loan applications, and decreasing the probability of rejecting creditworthy applicants or accepting high-risk ones, which in turn, will result in lower financial risks and higher profits. As a result, in credit scoring literature, it is well-established that a minor improvement in the performance of credit scoring models can result in significant savings \citep{baesens2003benchmarking}.

Over the years, several events have stressed the benefits and the need to develop more effective credit scoring models. For instance, one of the factors that resulted in the financial crisis of 2008 was the massive default of subprime borrowers in mortgage markets, which ultimately triggered the collapse of many financial institutes. To respond to this crisis, financial institutes have put a concerted effort in improving their measures in managing their financial risks by developing more effective credit scoring models \citep{garcia2019exploring}. In addition, with the emergence of digital technology, the financial sector has dramatically changed. Recently, Peer-to-Peer lending online platforms have emerged as an alternative to traditional lending markets that allow individuals to directly borrow money from lenders without the financial institutes’ intermediations. Unlike traditional lending, the offered loans in peer-to-peer platforms are not secured and cannot be backed up by collateral. Hence, it is crucial to determine the default risk associated with granting a loan to each borrower \citep{emekter2015evaluating,bastani2019wide}. Moreover, due to digitalization and new alternative lending markets, such as peer-to-peer lending, loan portfolios have expanded. Therefore, increasing customer data, such as information on their credit history, demographic, and financial data, is being accumulated. This data can be used to propose and implement credit scoring models based on machine learning techniques to reduce financial risk resulting from loan default. Accordingly, numerous credit scoring models based on various binary classification algorithms have been implemented to leverage the massive amounts of customers data throughout the years. For instance, Linear Discriminant Analysis (LDA) \citep{reichert1983examination}, Logistic Regression  \citep{serrano2015determinants}, Support Vector Machines (SVM) \citep{yu2018dbn}, and Random Forests \citep{malekipirbazari2015risk} are among the utilized classification techniques.

Retrospective studies in credit scoring have proposed several promising approaches to improve the classification performance in classifying loan applicants to help the decision-making process of banks and financial institutes. For example, ensemble methods that combine the decisions of several classification models have resulted in superior classification performance compared to individual models and therefore are widely used in credit scoring \citep{galar2011review,lessmann2015benchmarking,dastile2020statistical}. Profit-based credit scoring approaches that incorporate the profitability of loans in the classification process to maximize the obtained profits from the accepted and rejected loans have also been proposed \citep{bastani2019wide, kozodoi2019multi}. However, several challenges remain to be addressed in improving loan default classification performance. First, credit scoring datasets usually suffer from the class imbalance problem since the majority of instances in the dataset are customers who have been able to pay off their loans (negative class) \citep{brown2012experimental, xiao2012dynamic}. The class imbalance problem may adversely affect classification models’ generalization power since most of these models have been formulated to minimize an accuracy-based objective function in the training phase \citep{haixiang2017learning}. To alleviate this issue, several approaches have been developed that can be categorized as: Data level, algorithmic level, and cost-sensitive approaches, which are described in detail in Section \ref{imbalance}. Second, in general, credit scoring data are typically comprised of structured, semi-structured, and non-structured data that illustrate each loan applicant's demographic, financial, and behavioral information. In order to develop robust credit scoring classification techniques that perform well in practice, it is crucial to embed the underlying information from each aspect of the credit scoring dataset in a lower-dimensional representation of the input data \citep{lei2019generative, MALDONADO2017656}. Representation learning is an approach that derives lower-dimensional embeddings of initial input features to obtain a more expressive data representation that increases the generalization power \citep{najafabadi2015deep}. Among representation learning approaches, autoencoders have recently attracted much attention in extracting the input data's underlying patterns before classification \citep{bengio2012unsupervised}, which are used in this study.  Third, despite the superior performance of neural networks and deep learning algorithms in many research fields, these learning models have not yet been extensively applied in credit scoring. Thus, further exploration of neural network-based classifiers in credit scoring is required \citep{dastile2020statistical,bhatore2020machine,shen2020new}. 

Although several studies in credit scoring literature have focused on alleviating the negative impacts of class imbalance on credit scoring models, and some studies have utilized representation learning as a means to improve the classification performance, to the best of our knowledge, only one paper (\citep{wong2020cost}) has proposed a model that benefits from representation learning and addresses the adverse effect of imbalanced data simultaneously by developing a cost-sensitive classification model based on stacked autoencoders for business problems. Since a further investigation is needed to determine the power of representation learning-based classifiers in imbalanced environments, we propose the Bagging Supervised Autoencoder Classifier in this paper.

In this paper, we propose the Bagging Supervised Autoencoder Classifier (BSAC) to tackle the challenges of class imbalance and improve classification generalization by representation learning. BSAC benefits from the representation power of the Supervised Autoencoders, which enhances the power of autoencoders in learning new representations by applying the principles of multi-task learning in order to extract better representations that are guided to learn new embeddings exclusively with regards to the final classification task. In addition, BSAC employs under-sampling and ensemble learning, which are subcategories of data level and algorithmic level approaches in handling imbalanced datasets, respectively, to further improve the classification performance in the imbalanced setting of credit scoring. The main contributions of this paper are:

\begin{itemize}
\item To find the optimal representation of credit scoring input data, we use the supervised autoencoder, which is based on the principles of multi-task learning, in learning a lower-dimensional embedding guided by the final classification task (Section \ref{SA_desc}). 
\item To decrease the negative effects of class imbalance in credit scoring, we integrate data level and algorithmic level approaches in dealing with imbalanced data, namely a variant of the bagging process and under-sampling of the majority data, respectively (Section \ref{subset_create}) and evaluate the effectiveness and superiority of obtained results from BSAC against previous papers..  
\end{itemize}

\section{Related Works}
\subsection{Data Imbalance}\label{imbalance}
In many real-life classification problems, the number of observations belonging to the positive class is significantly smaller than the number of observed negative instances. This class imbalance problem may adversely affect the machine learning classification algorithms' performance, as most of them minimize an accuracy-based objective function in the training phase. Therefore, the classification results on an imbalanced dataset may be biased towards the majority class \citep{haixiang2017learning}. 

So far, much research have been focused on minimizing the effect of class imbalance on the final predictions of learning algorithms in credit scoring \citep{brown2012experimental, yu2018dbn, he2018novel}. Most of these efforts can be classified into three categories: Data level, Algorithm level, and Cost-sensitive algorithms \citep{galar2011review, haixiang2017learning}. It is worth mentioning that the following categories overlap in most works and cannot be classified as belonging to only one of these categories. 

\subsubsection{Data level}
The imbalance ratio of a dataset is defined as the number of samples with negative class (majority class) divided by the number of samples with positive class (minority class). 

Data level approaches in imbalanced learning intend to balance out the class distribution by resampling methods such that the imbalance ratio of the dataset is approximately close to 1. Random Undersampling (RUS), random Oversampling (ROS), Synthetic Minority Oversampling Technique (SMOTE), and its variants are among these methods. 

Random undersampling, randomly selects a number of observations in the majority sample, and discards them until the number of majority and minority class samples are roughly the same. One of the major drawbacks of RUS is that valuable information may be lost during the process. Random oversampling, randomly selects and replicates a sample from the minority class to modify the class distribution in the data set to obtain an approximate imbalance raio of one. 

Since directly oversampling the minority class instances may lead to overfitting and lack of generalization, \citet{chawla2002smote} proposed synthetic  minority  oversampling  technique to overcome this problem. In their approach, instead of oversampling minority samples, synthetic samples are generated by computing the distance between a minority class sample and its nearest neighbor, and multiplying this distance by a random number between 0 and 1 and add it to the minority sample's feature vector. This will generate a synthetic sample that belongs to the minority class, which leads to a more general decision region of the minority class. 

Resampling methods have been implemented in several studies in credit scoring literature.  For instance, \citet{brown2012experimental} conducted a study to examine the robustness of several classification methods in classifying imbalanced credit scoring datasets. They increased the class imbalance severity of their data by undersampling the positive class to observe the changes in classification performance in environments with higher imbalance ratio. 
\citet{he2018novel} extend the Balance Cascade technique, which is a supervised undersampling method to create data subsets as inputs to an ensemble learning model. They evaluated their ensemble model by applying it to several credit scoring datasets. \citet{papouskova2019two} developed a two-stage ensemble learning model using EUSBoost, which is an evolutionary undersampling method to overcome the problems of class imbalance in credit scoring. 

Regardless of implemented classification models and tenchinques, several studies \citep{duan2019financial, shen2020new, chen2019credit} employed SMOTE or a variation of it as a data pre-processing step to enhance the robustness of classification algorithms in classifying highly imbalanced data sets.  

\subsubsection{Cost-sensitive learning}
In practical applications of machine learning, the misclassification costs of different classes are not equal. For instance, it is stated that misclassifying a defaulted loan as a non-default application is more costly for financial institutions \citep{garcia2019exploring}. Many classification algorithms assume an equal misclassification cost for minority and majority classes in imbalanced data sets. Cost-sensitive approaches try to incorporate misclassification costs into the classifiers' training phase to improve the classification performance of minority samples. The misclassification cost can be represented as $c_{ij}$, which is the cost of misclassifying a sample as class $i$ instead of class $j$. 

In credit scoring literature, many researchers have focused on developing models that incorporate misclassifcation costs into the original cost function of classifiers.  \citet{xiao2012dynamic} proposed a cost-sensitive dynamic ensemble method that dynamically selects classifiers for each test sample classification. \citet{feng2018dynamic} also developed a dynamic ensemble classification model for credit scoring based on the relative costs of Type I and Type II error which selects a subset of classifiers for classifying test samples.
\citet{xia2017cost} propose a cost-sensitive boosted tree loan evaluation model to enhance the classification performance in identifying the minority class instances from majority class ones.
In their proposed cost-sensitive model, \citet{wong2020cost} assigned a higher misclassification cost to the minority class instances in an ensemble of stacked autoencoders to increase the classification capability. Several other research papers such as \citep{bahnsen2015example, xiao2020cost} have also proposed different classification models based on cost-sensitive learning in classification of defaulted and non-default loans.

\subsubsection{Algorithmic level}\label{Alg_lvl_lit}
Algorithmic level approaches attempt to modify algorithms to improve their performance in an imbalanced setting. For instance, modified Support Vector Machine (SVM) \citep{batuwita2010fsvm, chen2019credit}, logistic regression \citep{maalouf2011robust}, and several other studies have been proposed in this category. For a detailed review of other modified algorithms, we refer to \citet{haixiang2017learning} review of imbalanced learning.

The second approach of the algorithmic level is the ensemble learning method. Ensemble learning models, also known as multiple classifier systems, aggregate the decisions of different accurate and diverse individual (base) classifiers. Ensemble learning methods were developed under the motivation that different, diverse classifiers often make different mistakes because of their innate distinctions in classifying the same data. Therefore, if these classifiers' decisions are aggregated together, they can compensate for each others' mistakes when making the final prediction \citep{kittler1998combining}.

Empirical research has shown that ensemble models have superior performance in credit scoring\citep{lessmann2015benchmarking, wang2011comparative}. In addition, out of 527 papers reviewed by \citet{haixiang2017learning}, 218 papers proposed novel ensemble models in various imbalanced classification tasks. Specifically, in credit scoring, the majority of existing literature revolves around developing ensemble classification models since \citet{lessmann2015benchmarking} and \citet{wang2011comparative} have demonstrated that compared to other classification algorithms in credit scoring, ensemble classifiers show an outstanding classification performance.

As a result of ensemble models' superior performance in credit scoring, \citet{yu2018dbn} used support vector machines as the base classifier of their ensemble model. After applying a re-sampling method to generate balanced training subsets, SVM classifiers are trained. To reach the final prediction of their proposed model, they use a deep belief network to fuse the predictions of base classifiers. 
\citet{xia2018novel} also uses a trainable classifier, namely extreme gradient boosting algorithm, to fuse the prediction of base classifiers. Their proposed heterogeneous ensemble model (i.e., an ensemble model with different base classifiers) integrates the bagging algorithm with stacking. \citet{sun2018imbalanced} proposed an ensemble of decision tree classifiers to classify the imbalanced data of enterprise credit evaluation. They integrated bagging, SMOTE and differentiated sampling rates to create diverse subsets of data as as input to decision tree classifiers. Since different subsets of data were created, the final trained decision trees will be diverse classifiers that can perform better in an imbalanced environment. 
So far, much research has been conducted that integrate the data level, cost-sensitive learning, and algorithmic level to explore the improvements that can be made in credit scoring classification \citep{he2018novel, garcia2019exploring, xiao2016ensemble}.

As a result of Ensemble learning superior performance in imbalanced data sets, in this study, we employ ensemble learning as well as re-sampling methods to classify a highly imbalanced real-life credit scoring dataset.

\subsection{Neural Networks}
An Artificial Neural Network (ANN) is a mathematical model inspired by the human brain's biological neural structure. ANNs attempt to imitate the way the biological neural structures process information by a network of interconnected neurons. Artificial Neural Networks usually consist of three layers: an input layer, a hidden layer, and an output layer. Each layer in ANN contains several neurons that are connected to the neurons of the consequent layer by weighted connections. During the training phase of a classification problem, the neural network's weights are adjusted by the backpropagation process, such that the error (usually Mean Squared Error (MSE)) between the ANN's output and the actual label of the training samples is minimized. A more comprehensive introduction to Artificial Neural Networks is presented in Section \ref{ann}.

Variants of Artificial Neural Networks have been used in the credit scoring literature. \citet{bastani2019wide} use the wide and deep model proposed by Google to develop a two stage credit scoring model that identifies most profitable loans for investment. In their proposed model, the first stage is responsible to identify non-default loans. In the second stage, the profitability of the identified non-default loans is predicted. In their work, \citet{shen2020new} improved the synthetic minority oversampling technique (SMOTE) along with a deep ensemble learning model to increase the classification performance of credit risk evaluation. \citet{neagoe2018deep} compared the performance of deep convolutional neural networks with multi-layer perceptron in the credit scoring classification problem. The result of their study indicates that the deep convolutional neural networks improve the overall accuracy of classification significantly. 

Although, studies have been conducted by implementing Artificial Neural Networks and deep learning approaches in the credit scoring literature \citep{yu2018dbn, duan2019financial}, they are still among the least frequently used classifiers in this area of research and further exploration of these models in credit scoring in required \citep{dastile2020statistical,shen2020new, bhatore2020machine}. 

\subsubsection{Autoencoders}
An Autoencoder is a three-layer neural network with an input, hidden, and an output layer that attempts to reconstruct the input data as its output. In the training phase, the input variables are the same as the target variables and the parameters of the Autoencoder, specifically its weights, are adjusted to minimize the error between the actual input data and the reconstruction of the training set. The final goal of training Autoencoders is to learn the useful representation of the input data in the hidden layer, which enables the Autoencoder to reconstruct the input data as the output \citep{Goodfellow-et-al-2016}. To increase the generalization power of other machine learning tasks, such as classification, the hidden representations of Autoencoders can be used as the input of other machine learning models. a comprehensive introduction to Autoencoders is presented in Section \ref{ae}.

In the credit scoring literature, \citet{tran2016credit} use genetic programming and stacked Autoencoders to develop a two-step hybrid credit scoring model. In the training phase, if-then classification rules are extracted from the training data by genetic programming and the stacked Autoencoder is trained. In the testing phase, first the test data is classified by the extracted rules. The stacked Autoencoder will classify the remainder of data that could not be classified by the extracted rules. \citet{wong2020cost} also used stacked Autoencoders to classify six datasets in the business domain. They proposed the Cost-Sensitive Deep Neural Network (CSDNN) and the Cost-Sensitive Deep Neural Network Ensemble (CSDE) to address the common class imbalance problem in the business domain datasets such as direct marketing, credit scoring, and fraud detection. They modified the Stacked Denoising Autoencoders (SDA) cost function by incorporating the majority and minority class's misclassification cost vector into the original cost function.

\citet{fan2018denoising} used Denoising Autoencoders to reduce noise in the data, to improve the classification accuracy of the Lending Club dataset. A multi-stage ensemble learning model was proposed by \citet{chen2019credit} for credit risk prediction. Gradient boosting decision trees and Autoencoders were used to generate new features and dimensionality reduction, respectively. Furthermore, logistic regression was applied to the extracted features from the previous stage to classify borrowers' credit risk. 
\citet{mancisidor2018segment} create customer clusters to identify segments that have different propensities of default on their loans. First, they transform the input data by the Weight of Evidence transformation, based on the exceeded days past due date to encapsulate the probability of default. Then, they used the Variational Autoencoder latent space, where the data's complex and non-linear relationships exist for clustering customers in segments with different propensities of default to finalize the classification.  

Based on the reviewed credit scoring approaches, numerous studies in this field focus on the class imbalance problem, and few address the use of neural networks and representation learning in improving the classification performance. To the best of our knowledge, only one paper has utilized neural networks, representation learning and approaches in mitigating the effects of class imbalance problem in credit scoring. Therefore, to address this shortcoming in credit scoring literature, in this study, an Autoencoder-based classification system based on multi-task learning is proposed as a credit scoring model. The Bagging Supervised Autoencoder classifier leverages the potentials of ensemble learning, multi-task learning, representation learning to handle imbalanced credit scoring datasets. The BSAC is described in detail in Section \ref{methodology}.

\section{Preliminaries}
In this Section, we briefly introduce the related algorithms to this study such as, artificial neural networks, and Autoencoders which are adopted in our study. 

\subsection{Artificial Neural Networks}\label{ann}
An artificial neural network is a biology-inspired mathematical model that tries to emulate how a biological neural network processes information. Artificial neural networks typically consist of three layers: an input, hidden, and an output layer. Each layer is constructed by several units called neurons which are connected to the neurons of the subsequent layer with an associated weight. The output of each neuron is the sum product of its inputs with the corresponding associated weight. Then a non-linear activation function is passed to the output of each neuron to generate an activation output which is then passed to the neurons of the subsequent layer as an input. The activation output layer $l$ is denoted by $a^{l}$ which is defined in equation \ref{activation} and \ref{sumproduct}:

\begin{flalign}\label{activation}
   a^{l} = \alpha(z^{l})
\end{flalign}

\begin{flalign}\label{sumproduct}
    z^{l} = W^{l}a^{l-1}+b^{l}
\end{flalign}

Where $\alpha$ is the activation function, $z^{l}$ is the linear output of layer $l$, $W^{l}$ is the weight matrix associated with the connection of neurons in layer $l-1$ and $l$, and $b^{l}$ is the bias vector of neurons in layer $l$.
During each iteration of the training phase, the weight matrix $W$ and the bias vector $b$ are modified and updated to minimize the cost function $J$ based on the comparison of the predicted label ($\hat{y}_i$) and the ground truth ($y_i$) by \textit{backpropagation} process. The Mean Squared Error (MSE, equation \ref{mse}) or in case of a binary classification problem binary cross entropy (equation \ref{bin_cross}) cost functions are usually used in the training process which for one training sample $x_i$ is calculated as follows.  

\begin{flalign}\label{mse}
    J_{MSE}=\frac{1}{N}\sum_{i=1}^{N}(y_i - \hat{y}_i)
\end{flalign}

\begin{flalign}\label{bin_cross}
    J_{CrossEntropy}=-\frac{1}{N}\sum_{i=1}^{N}y_i\cdot\log(p(y_i))+(1-y_i)\cdot\log(1-p(y_i))
\end{flalign}

\subsection{Autoencoders}\label{ae}
An Autoencoder is a three-layered artificial neural network structure that tries to reconstruct the input data in the output layer \citep{Goodfellow-et-al-2016}. The Autoencoder has a hidden layer that provides the representation or code that is utilized to reconstruct the input data in the output layer. The Autoencoder's structure can be described in two steps. First, an encoder function $h=f(x)$, encodes the input data into the hidden layer representation, then the decoder function $r=g(x)$ attempts to reconstruct the input data from the Autoencoder's hidden layer representation.

\begin{flalign}\label{ae_enc}
    h(x)=\alpha(Wx+b)
\end{flalign}

\begin{flalign}\label{ae_dec}
    r(h(x))=\alpha(W'h(x)+b')
\end{flalign}
In equation \ref{ae_enc} and \ref{ae_dec} the $W$ and $W'$ are the encoder and decoder weight matrix, respectively. Also $b$ and $b'$ are the bias vectors of encoder and decoder layers of the Autoencoder. Similar to the simple artificial neural networks, the weight matrix and the bias vector are optimized based on a cost function (equation \ref{auto_loss}) using the backpropagation process. In the Autoencoders cost function, the reconstruction error, which is the difference between the input data ($x_i$) and its reconstruction ($\hat{x}_i$) by the Autoencoder, is minimized as follows:

\begin{flalign}\label{auto_loss}
    L(x, \hat{x}) = \frac{1}{2}\sum_{i=1}{N}(\hat{x}_i-x_i)^2
\end{flalign}

\begin{figure}[h]
\centering
  \includegraphics[width=180pt]{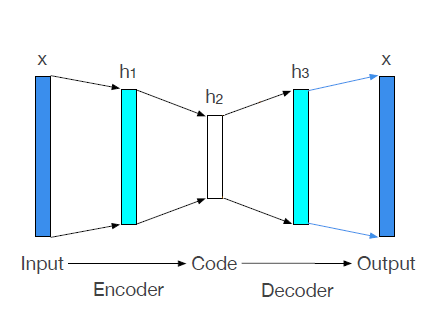}
  \caption{The general structure of an Autoencoder adopted from \citet{le2018supervised}. In the Autoencoder, the input data are first encoded to a new representation. Then, in the decoding phase, the network attempts to reconstruct the input data in the output layer using the new representation.}
  \label{fig:AE}
\end{figure}

The general structure of Autoencoders is depicted in figure \ref{fig:AE}.

\subsection{Multi-task Learning}\label{multi-decs}
Multi-task learning is a promising approach in improving the generalization of machine learning models by leveraging the available information that can be shared between other different, related learning tasks so that what is learned for one task can help improve the performance of other learning problems \citep{caruana1997multitask}.Multi-task learning can be observed in human behavior when learning different activities related to each other by sharing gained knowledge from one activity to learn another. Therefore, similar to human behavior, learning multiple learning tasks simultaneously in the learning process can be beneficial \citep{zhang2018overview}. 

In general, the typical approach in machine learning problems is to first break up a large problem into different, related small sub-problems and learn them separately. However, it is stated that this approach would result in abandoning and neglecting the valuable information contained in other tasks in the learning process of a specific task. Therefore, it is possible to construct models based on the available information of other models in parallel \citep{waibel1989modularity}. As a result, multi-task learning can be defined as: 

Given $T$ learning tasks $T=\{t_1,t_2,\dots,t_T\}$, where all or a subset of $t_i$ tasks are related to each other and are not identical, multi-task learning attempts to improve the learning performance of a model for task $t_i$ by using the knowledge learned and contained in all or a subset of $T-1$  other learning tasks simultaneously. 

Based on the principles of multi-task learning, and the importance of an expressive data representation in improving the machine learning tasks, The BSAC is developed for the credit scoring problem at hand, which will be described in Section \ref{BSAC}.

\section{Methodology}\label{methodology}
\subsection{Supervised Autoencoder}\label{SA_desc}
In a supervised learning problem, the ultimate goal is to train a model on the training data $D=\{(x_1,y_1),(x_2,y_2),\dots,(x_m,y_m)\}$ to learn a mapping function between the input data $x_i$ and its target $y_i$ that performs well in classifying unseen data. 
Increasing the generalization power of machine learning models and, in turn, increasing their performance in the classification task is one of the essential objectives in this setting. As mentioned in Section \ref{intro}, constructing a better and more expressive representation of the input data before performing classification is among the approaches that are implemented to increase the classification performance \citep{najafabadi2015deep}. Among the various models that have been proposed to improve the representation of the input data in machine learning, much attention has been paid to Autoencoders recently \citep{bengio2012unsupervised}.

The Supervised Autoencoder (SA) has been introduced with the idea of utilizing the representation power of Autoencoders in a supervised setting, with regards to the principles of multi-task learning \citep{le2018supervised}. Essentially, regarding the principles of multi-task learning, the Supervised Autoencoder can be defined as below: 

Given two learning tasks of representation learning and classification ($T=\{t_1, t_2\}$) that are related to each other, a Supervised Autoencoder attempts to both improve the  generalization power of classification, and enhance the expressiveness of the learned representation by the Autoencoder, using the information available in learning both tasks simultaneously. This goal is achieved by the addition of a supervised loss to the representation layer of the Autoencoder.

Adding a supervised loss to the representation layer of an Autoencoder directs the learning towards representations that are more effective in the classification task. In contrast, if the training process in representation learning takes place solely based on the input data, the Autoencoder may be able to extract underlying information well-fitted to the input data, but the found solutions may not generalize well in the subsequent classification task. In other words, the Supervised Autoencoder combines the reconstruction loss of the Autoencdoer with the prediction loss of the classification task to improve the target prediction performance by using representations specific to classification \citep{le2018supervised}. 

\begin{figure}[h]
\centering
  \includegraphics[width=180pt]{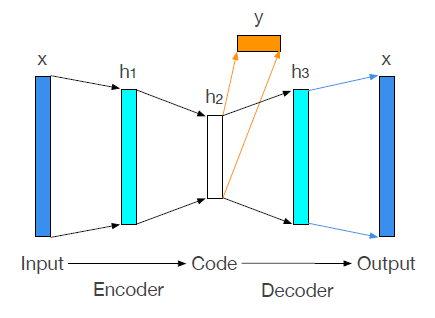}
  \caption{The general structure of a Supervised Autoencoder from \citet{le2018supervised}. The Supervised Autoencoder consists of adding a classification loss to the hidden (representation) layer of an Autoencoder to increase the classification performance by directing the new representations to fit better to the classification task.}
  \label{fig:SAE}
\end{figure}

In Supervised Autoencoder, the main objective is to create a network that learns a mapping function between the input data $D=\{(x_1,y_1),\dots,(x_m,y_m)\}$ to the output labels $y$ using the new representation that is simultaneously learned by the network in the learning process. In other words, the network consists of two main parts that jointly learn the new input data representation, and the final labels as shown in figure \ref{fig:SAE}. Therefore, the Supervised Autoencoder’s final cost function is defined as shown in equation \ref{SAE_eq}.

\begin{flalign}\label{SAE_eq}
\frac{1}{m}\sum_{i=1}^{m}[(\gamma)L_r(\hat{x_i},x_i)+(1-\gamma)L_p(\hat{y_i},y_i)]
\end{flalign}

Here, as shown in equation \ref{Lr}, $L_r$ is the reconstruction loss of the network, which is defined as a Mean Squared Error loss between the input data $x$ and the reconstructed values in the output layer of the Autoencoder $\hat{x}$.

\begin{flalign}\label{Lr}
L_r = \frac{1}{M}\sum_{i=1}^{N}(x - \hat{x})
\end{flalign}

The $L_p$ (equation \ref{Lp}) is the prediction loss that uses the learned representations to predict the labels of input data, which is defined as a binary cross entropy loss between the true labels $y_i$ of each data sample $x_i$ and the predicted labels $\hat{y_i}$.

\begin{flalign}\label{Lp}
L_p = -\frac{1}{M}\sum_{i=1}^{N}y_i\cdot\log(p(y_i))+(1-y_i)\cdot\log(1-p(y_i))
\end{flalign}

In equation \ref{SAE_eq}, the values of $\gamma$ and $1-\gamma$ are the weighting parameters of $L_r$ and $L_p$, respectively, which are incorporated in the cost function in order to assess and balance the contributions each term, namely the reconstruction loss and the prediction loss, in the final structure. The value of $\gamma$ is optimized by the grid search process using the validation set. It is worth to emphasize that the optimal weights of the connections between neurons of each layer is learned simultaneously by the reconstruction loss and prediction loss.

It has been shown, both empirically and theoretically, that Supervised Autoencoders can improve the generalization power of neural networks with the same structure (minus the Autoencoder) \citep{le2018supervised}. As a result, In this paper a Bagging Supervised Autoencoder Classifier is proposed in loan data classification

\subsection{Bagging Supervised Autoencoder Classifier}\label{BSAC}
This Section describes the general structure of the proposed Bagging Supervised Autoencoder Classifier.
The BSAC consists of three main steps described in the subsequent Sections:

\subsection{Step 1: Creating the subsets of data}\label{subset_create}
The first step in the BSAC is to create subsets of data based on the Bagging procedure. The bagging algorithm iss one of the most effective algorithms in ensemble learning due to its power in decreasing the variance and as a result, preventing overfitting \citep{breiman1996bagging}. The BSAC utilizes the principles of the Bagging process to decrease overfitting and increase the model’s robustness in handling credit scoring’s imbalanced data.

In order to handle the class imbalance problem in credit scoring more effectively, here, we use a variant of the Bagging procedure by incorporating subsampling into Bagging to train base classifiers (Supervised Autoencoders) on. 

Given a dataset with $m$ samples such that $m=p+n$, and $p$ and $n$ are the number of minority and majority samples respectively, balanced subsets of data are created. Specifically, first, the minority samples are set aside. Then, random subsampling is applies on the majority samples to select approximately $p$ negative samples from the customer data that have successfully paid off their loans. Then, the randomly selected majority samples are combined with the $p$ minority samples to create a balanced subset that consists of $2p$ samples. The number of created subsamples are determined by the rounding down imbalance ratio of the data set. Finally, all subsets are used as training data for Supervised Autoencoders, which are considered our base classifiers, here. 

It is worth mentioning that in the proposed model, valuable information will not be lost in  the sampling process, since all samples in the majority class will be used to train the base classifiers. In addition, the sampling method can prevent overfitting by training the base classifiers on balanced subsets of data.

\subsubsection{Step 2: Training base classifiers}
After creating random subsets of the data, each subset is utilized to train a Supervised Autoencoder to both reconstruct data representations, and classify each samples. In the training process, the optimal values of $\gamma$ and $1-\gamma$ are determined by using the validation set based on the F1 measure since both recall and precision are an integral part in an effective credit scoring system in distinguishing the differences between minority and majority samples, especially in an imbalanced setting. 

\subsubsection{Step 3: Base Classifiers Aggregation}
After training each Supervised Autoencoder as base classifiers and determining the optimal value for $\gamma$, the majority voting process is utilized to aggregate the results of individual classifier to reach to the final prediction. 

\begin{figure}[h]
\centering
  \includegraphics[width=230pt]{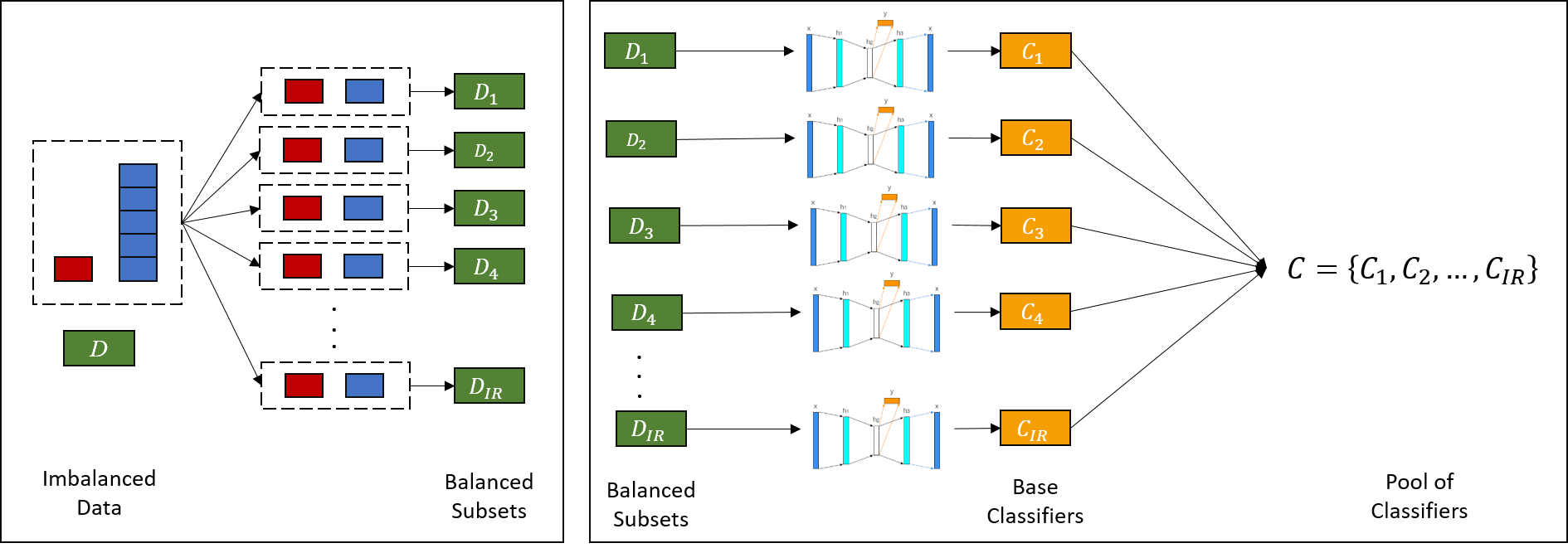}
  \caption{The training phase of the BSAC. In this model, first, balanced subsets of data are randomly created from the imbalanced training data. Then, each subset is used as the training set of the Supervised Autoencoder as the base classifier. Finally, the predictions of each base classifiers are aggregated by the majority voting process.}
  \label{fig:BSAC_train}
\end{figure}

\begin{figure}[h]
\centering
  \includegraphics[width=180pt]{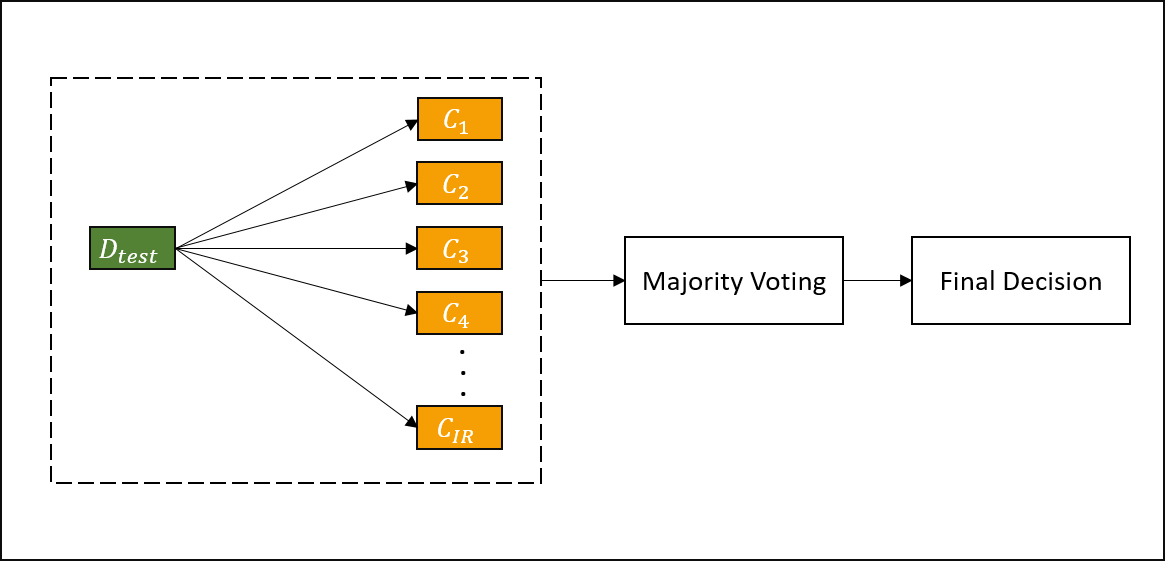}
  \caption{The testing phase of the BSAC. After training the Supervised Autoencoders as base classifiers using the balanced subsets, each test sample is labeled by the base classifiers, and the final output is determined by the majority voting process.}
  \label{fig:BSAC_test}
\end{figure}

As a result, the BSAC combines ensemble learning, representation learning, subsampling, and multi-task learning to effectively handle the imbalance data of credit scoring and improve the classification performance. Algorithm \ref{BSAC_alg} illustrates the classification process of the BSAC. In addition, figures \ref{fig:BSAC_train} and \ref{fig:BSAC_test} depict the general training and testing process of BSAC, respectively. 

\begin{algorithm}
\SetAlgoLined
	\KwIn{Data set with m samples, $D=\{(x_1,y_1),(x_2,y_2),\dots,(x_m,y_m)\}$\\
	$p$: The number of positive samples in $D$\\
	$n$: The number of negative samples in $D$\\
	$IR=\frac{n}{p}$: The imbalance ratio\\
	$SAE$: The supervised Autoencoder classifier\\
	$C = \emptyset$: The pool of trained classifiers}
	\For{$i \in IR$}{
	$p_i$: Choose $p$ samples randomly without replacement from the positive samples in the data set.\\
	$D_i = p_i \cup n$ The balanced data set used for training each $SAE$.\\
	$c_i = SAE(D_i)$: The trained supervised Autoencoder on the balanced data set.\\
	$C \cup c_i$: Update the pool of trained classifiers
	 }
	 \Return{The pool of classifiers $C=\{c_1,c_2,\dots,c_{IR}\}$}
	 \caption{The BSAC Algorithm}
	 \label{BSAC_alg}
\end{algorithm}

\section{Experimental Setting}
\subsection{Datasets}
\subsubsection{Taiwan Credit Data}
The Taiwan credit data~\footnote{retrieved from \href{https://archive.ics.uci.edu/ml/datasets/default+of+credit+card+clients}{https://archive.ics.uci.edu/ml/datasets/Taiwan}} is used in this study as a benchmarking dataset. The Taiwan credit dataset contains 30000 data samples with 24 variables to be used in a binary classification task to predict customers' loan default. After using dummy variables for categorical features the final dataset has 33 variables including the target value. The imbalance ratio of the Taiwan dataset is 3.52.

\subsubsection{Lending Club Dataset}
In this study, we also use the Lending Club dataset~\footnote{retrieved from \url{https://-www.kaggle.com/wordsforthewise/lending-club}}, which is one of the most popular online lending platforms in the financial industry. The loan data of the Landing Club dataset consists of loans with the payback duration of either $36$ or $60$ months. Since the latest payment made by a customer in our dataset was from March 2019, we selected $36$ month loans issued before March 2016 to ensure that the loan has had enough time to reach maturity. Out of the $151$ variables available in our dataset, we removed variables that included more than half the size of the dataset missing values. The remaining variables were organized such that variables without useful information, such as customers’ \textit{id} or \textit{URL}, were dropped. 

The \textit{average\_fico} was engineered from the average values of original variables \textit{fico\_range\_high}, and \textit{fico\_range\_low} for each data sample. We also created the \textit{credit\_history} variable by calculating the credit history of each customer from variables \textit{earliest\_cr\_line
} and \textit{issue\_d}. Finally, continuous variables were normalized, and as the last step in data preparation, we used dummy variables to  represent categorical features of our dataset. The final variables and their descriptions are included in Table \ref{LC_var} in appendix. The final dataset contains 206732 samples with 82 variables, including the target value of loan status. The imbalance ratio of our dataset was calculated as 5.98.

\subsection{Evaluation Metrics}
To evaluate the performance of the techniques studied in this paper, we consider Accuracy, Precision and Recall, F-measure, and G-mean as the evaluation metrics. The selected evaluation metrics have been chosen as they can cover different aspects regarding the performance of a classification model. 

\subsection{Experiments}
To evaluate the performance of our proposed method, BSAC, we use k-fold cross validation as follows. The value of $k$ was determined as $5$. Therefore, we split our dataset into $5$ equal folds and in each iteration, we hold out one fold as the test set, one fold as the validation set to optimize the hyper-parameter $\gamma$, and and use the remaining $5-1-1$ folds as training data. It is worth mentioning that each fold contains the same ratio of negative and positive samples to increase the training process robustness against the class imbalance in our data set. 

First, we evaluate the performance of our model using the Taiwan Credit data, which is used as the benchmark to compare our results with previous studies in this field. The structure of our Bagging Supervised Autoencoder was defined as \{32, 16, 8, 5, 8, 16, 32\} so that each layer contains approximately half the number of neurons from the previous layer. Our experiments were run for 200 epochs. 

Second, after verifying the performance of our model on the benchmark data, we apply the BSAC on the Lending Club dataset. Similar to the Taiwan experiments, we use 5-fold cross validation by splitting the data into $5$ equal folds and using one fold as the test set, and one fold from the training data as the validation set to optimize hyper-parameter $\gamma$. The structure of the BSAC is defined as \{81, 60, 30, 15, 30, 60, 81\}, which is run for 200 epochs on our dataset. 

During the validation process, the hyper-parameter $\gamma$ is optimized using the validation fold, by searching for the best value of $\gamma$ in the search space of $\gamma \in \{0.9, 0.8, \dots, 0.1\}$. 
The experiments were implemented with python, keras and tensorflow 2.0. 

\subsubsection{Hyper-parameter Selection and Classifier Diversity}
As mentioned in Section \ref{Alg_lvl_lit}, ensemble classification models benefit from both the accuracy and the diversity of base classifiers in the pool of classifiers to increase the classification performance. Several approaches can be implemented to increase the diversity of classifiers used in an ensemble learning model \citep{duin2002combining}. The following approaches are ordered based on the level of diversity from lower to higher diversities:

\begin{enumerate}
    \item \textbf{Different initialization}: Such as changing the initial configuration of a neural networks.
    \item \textbf{Different parameters:} Such as changing the kernels of a Support Vector Machine.
    \item \textbf{Different architecture:} Such as changing the structure and the number of layers in a neural network.
    \item \textbf{Different classifiers:} Such as utilizing a heterogeneous pool of classifiers.
    \item \textbf{Different training sets:} Such as using different training data for each base classifier of the pool.
    \item \textbf{Different feature sets:} Such as using different features for each base classifier of the pool.
\end{enumerate}

In the Bagging Supervised Autoencoder Classifier proposed in this paper, two approaches have been used in order to increase the diversity of the base classifiers. First, we use different subsets of balanced data, $D_i = p_i \cup n$, which is comprised of the same number of positive and negative samples to train the base classifiers. In addition to using different training sets for each base classifiers, we also tune the parameters of each base classifier separately with regards to its performance on the validation data. Therefore, each base classifier in the final ensemble pool of classifiers uses different hyper-parameters, which guarantees diversity to some extent. 

\section{Results and Discussion}
\subsection{BSAC performance}
The obtained results from the experiments for the Taiwan credit data is illustrated in Table \ref{Taiwan_res}. The mean and standard deviation of each evaluation metric is reported in the last two rows of Table \ref{Taiwan_res} in order to depict the stability of the Bagging Supervised Autoencoder. 

\begin{table}[h]
\centering
\caption{Obtained results from the BSAC model on Taiwan credit dataset after optimizing hyper-parameter $\gamma$}
\begin{tabular}{lcccc}
\hline
Fold & Recall (TPR) & F1 score    & G-mean      & Specificity (TNR) \\ \hline
0    & 0.5783       & 0.5426      & 0.6981      & 0.8427            \\
1    & 0.4861       & 0.5200        & 0.6581      & 0.8911            \\
2    & 0.5463       & 0.5286      & 0.6823      & 0.8521            \\
3    & 0.5735       & 0.5371      & 0.6942      & 0.8404            \\
4    & 0.5109       & 0.5297      & 0.671       & 0.8812            \\ \hline
mean & 0.5390      & 0.5316     & 0.6807     & 0.8615            \\
std  & 0.0357  & 0.0077 & 0.0147 & 0.0207       \\ \hline
\end{tabular}
\label{Taiwan_res}
\end{table}

The Taiwan credit data has been used in this study to verify the performance of our proposed model against other models in the literature. We have selected three papers that address the imbalanced nature of the credit data using either ensemble learning, sampling, or by implementing an Autoencoder-based model. The comparison between the results of our model and the selected papers are illustrated in Table \ref{comp_bench}. The comparison of the results of BSAC across different evaluation metrics to other studies that have been conducted using the Taiwan dataset, indicates that BSAC is capable of classifying credit scoring data effectively. It is worth mentioning that the obtained competitive results are only based on optimizing hyper-parameter $\gamma$.

\begin{table}[h]
\centering
\caption{Comparison of BSAC results with similar papers in the literature.}
\resizebox{\textwidth}{!}{%
\begin{tabular}{lcccc}
\hline
\multicolumn{1}{c}{\multirow{2}{*}{Classification   Model}} & \multicolumn{4}{c}{Evaluation   Metric}                   \\ \cline{2-5} 
\multicolumn{1}{c}{}                                        & Recall       & Specificity  & F1 score     & G-mean       \\ \cline{2-5} 
\citet{garcia2019exploring}                                       & 0.396        & 0.961        & not reported & not reported \\
\citet{he2018novel}                                       & not reported & not reported & 0.50415      & 0.62006      \\
\citet{wong2020cost}                                       & 0.62         & 0.78         & not reported & 0.7          \\
BSAC (proposed model)                                       & 0.5390       & 0.8615       & 0.5316       & 0.6807       \\ \hline
\end{tabular}
}
\label{comp_bench}
\end{table}

Table \ref{LC_res} shows the obtained results from the experiments on the Lending Club dataset, which was used as a real-life data set to illustrate the performance of our model. 

\begin{table}[h]
\centering
\caption{Obtained results from the BSAC model on Lending Club dataset after optimizing hyper-parameter $\gamma$}
\begin{tabular}{lcccc}
\hline
Fold   & Recall (TPR) & F1 score    & G-mean      & Specificity (TNR) \\ \hline
0       & 0.5572       & 0.3400        & 0.6292      & 0.7104            \\
1       & 0.5425       & 0.3356      & 0.6231      & 0.7156            \\
2       & 0.5560        & 0.3369      & 0.6267      & 0.7065            \\
3       & 0.5317       & 0.3346      & 0.6200        & 0.7230             \\
4       & 0.5465       & 0.3438      & 0.6296      & 0.7252            \\ \hline
mean    & 0.5468      & 0.3382     & 0.6257     & 0.7161           \\
std     & 0.0094  & 0.0033 & 0.0037 & 0.0071       \\\hline
\end{tabular}
\label{LC_res}
\end{table}
\subsection{Sensitivity Analysis}
As mentioned in Section \ref{methodology}, the loss function of BSAC is comprised of two main terms, namely the reconstruction loss $L_r$ and the prediction loss $L_p$. The hyper-parameter $\gamma$ controls the contribution of $L_r$ and $L_p$ in the final classification model, which is selected by hyper-parameter optimization during the training phase. 

Therefore, after the validation and hyper-parameter selection process, a high value of $\gamma$ indicates that the final outcome of the classification task, is highly dependent on the new representation of the input data by reconstructing data samples. Similarly, a low $\gamma$ value decreases the dependency of BSAC to the reconstruction loss and a neural network that has the same structure as the encoder part of the Autoencoder would have been sufficient for obtaining a high classification performance.

\begin{figure}[ht]
  \centering
  \begin{subfigure}[b]{0.3\linewidth}
    \includegraphics[width=\linewidth]{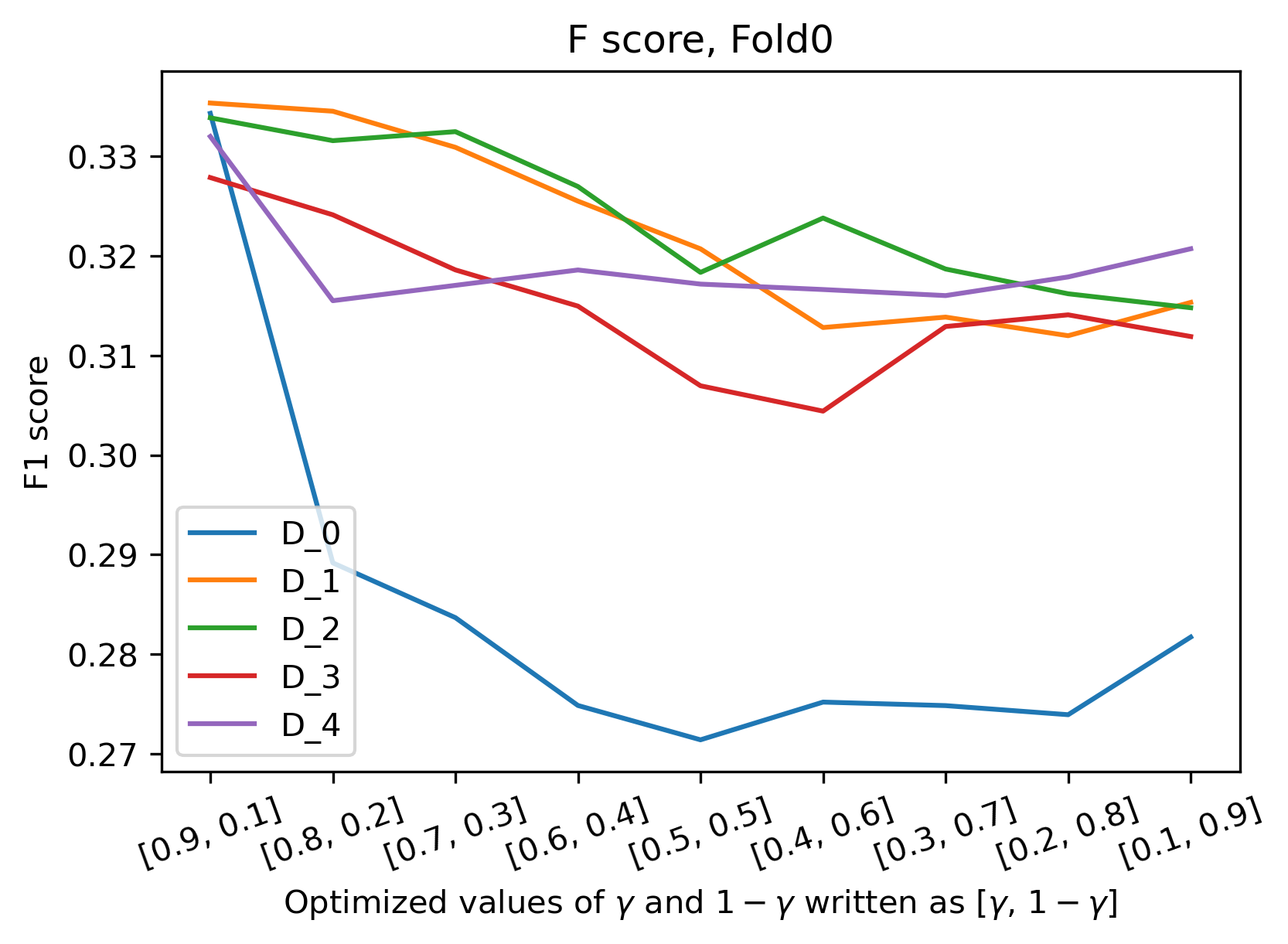}
    \caption{\tiny The changes in the F1 measure of validation set in Fold 0.}
  \end{subfigure}
  \begin{subfigure}[b]{0.3\linewidth}
    \includegraphics[width=\linewidth]{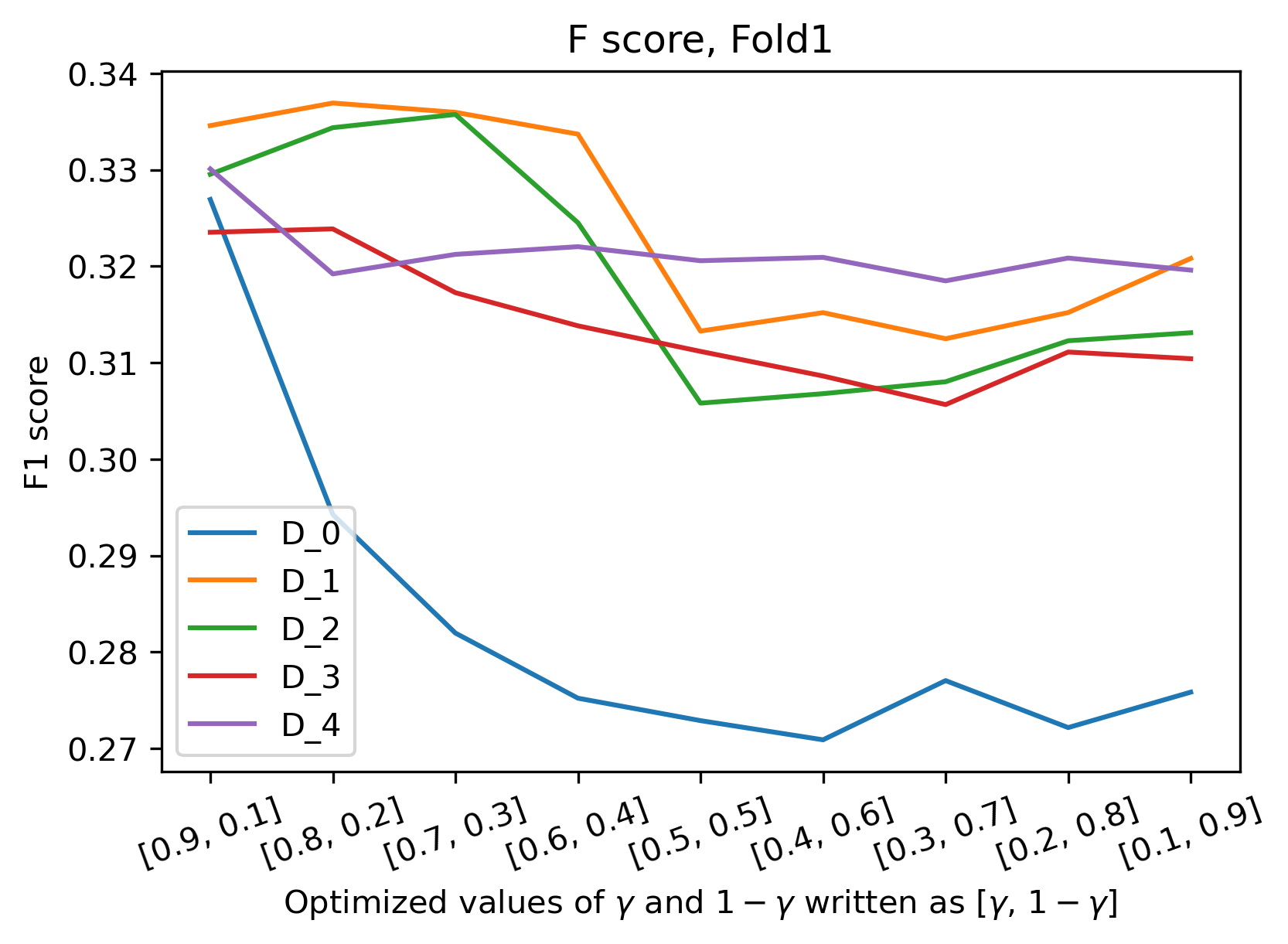}
    \caption{\tiny The changes in the F1 measure of validation set in Fold 1.}
  \end{subfigure}
  \begin{subfigure}[b]{0.3\linewidth}
    \includegraphics[width=\linewidth]{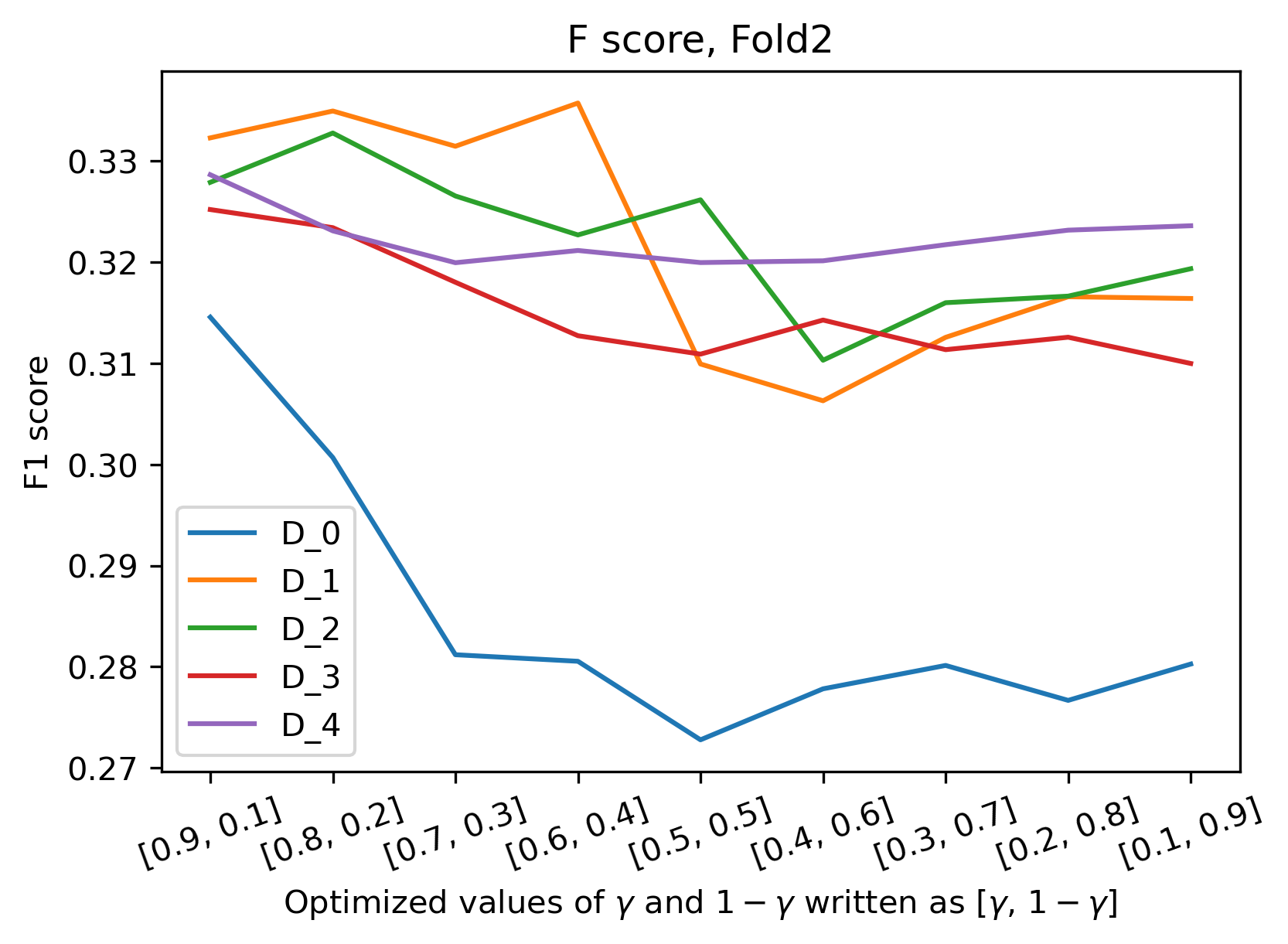}
    \caption{\tiny The changes in the F1 measure of validation set in Fold 2.}
  \end{subfigure}
  \begin{subfigure}[b]{0.3\linewidth}
    \includegraphics[width=\linewidth]{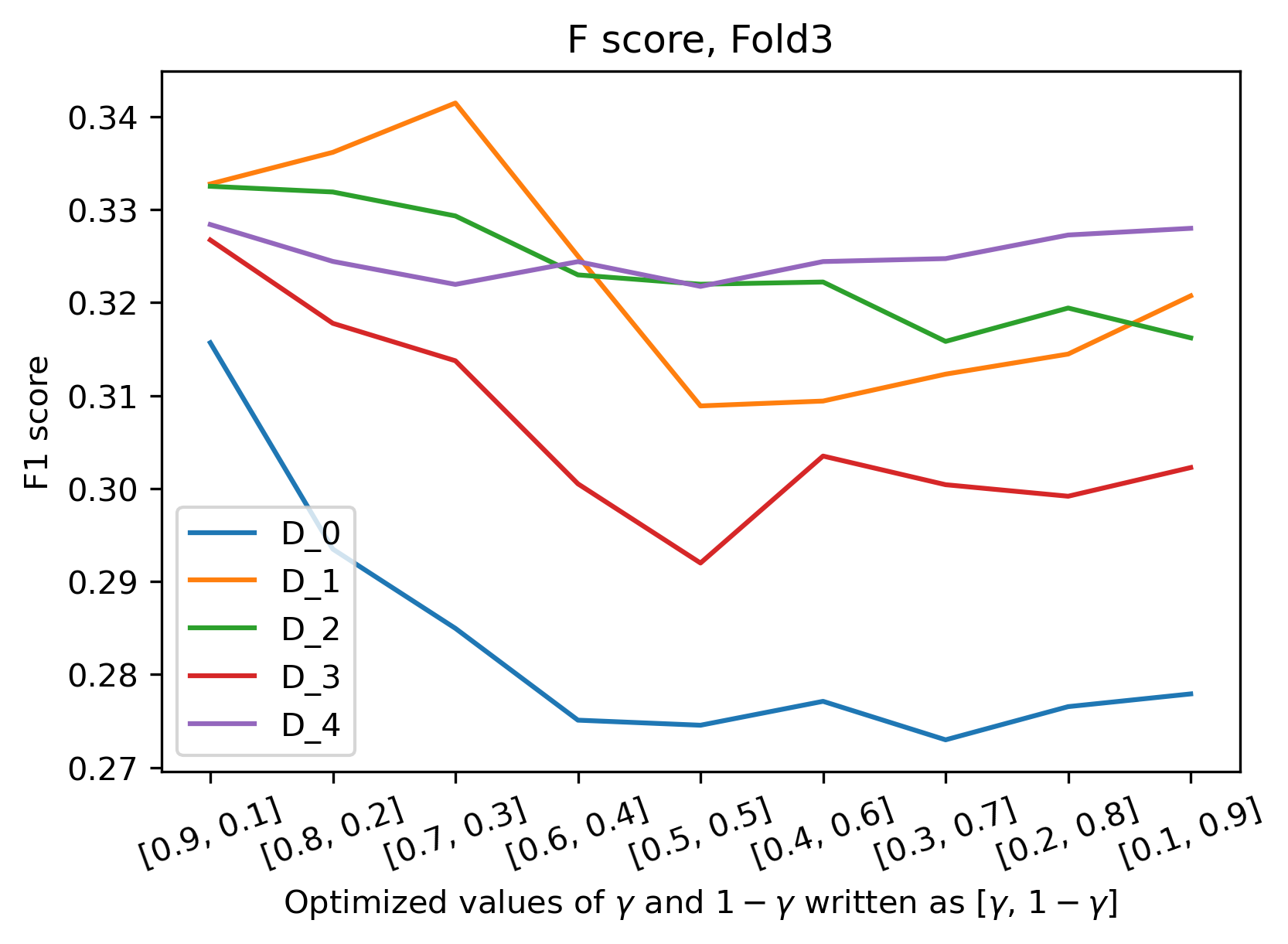}
    \caption{\tiny The changes in the F1 measure of validation set in Fold 3.}
  \end{subfigure}
  \begin{subfigure}[b]{0.3\linewidth}
    \includegraphics[width=\linewidth]{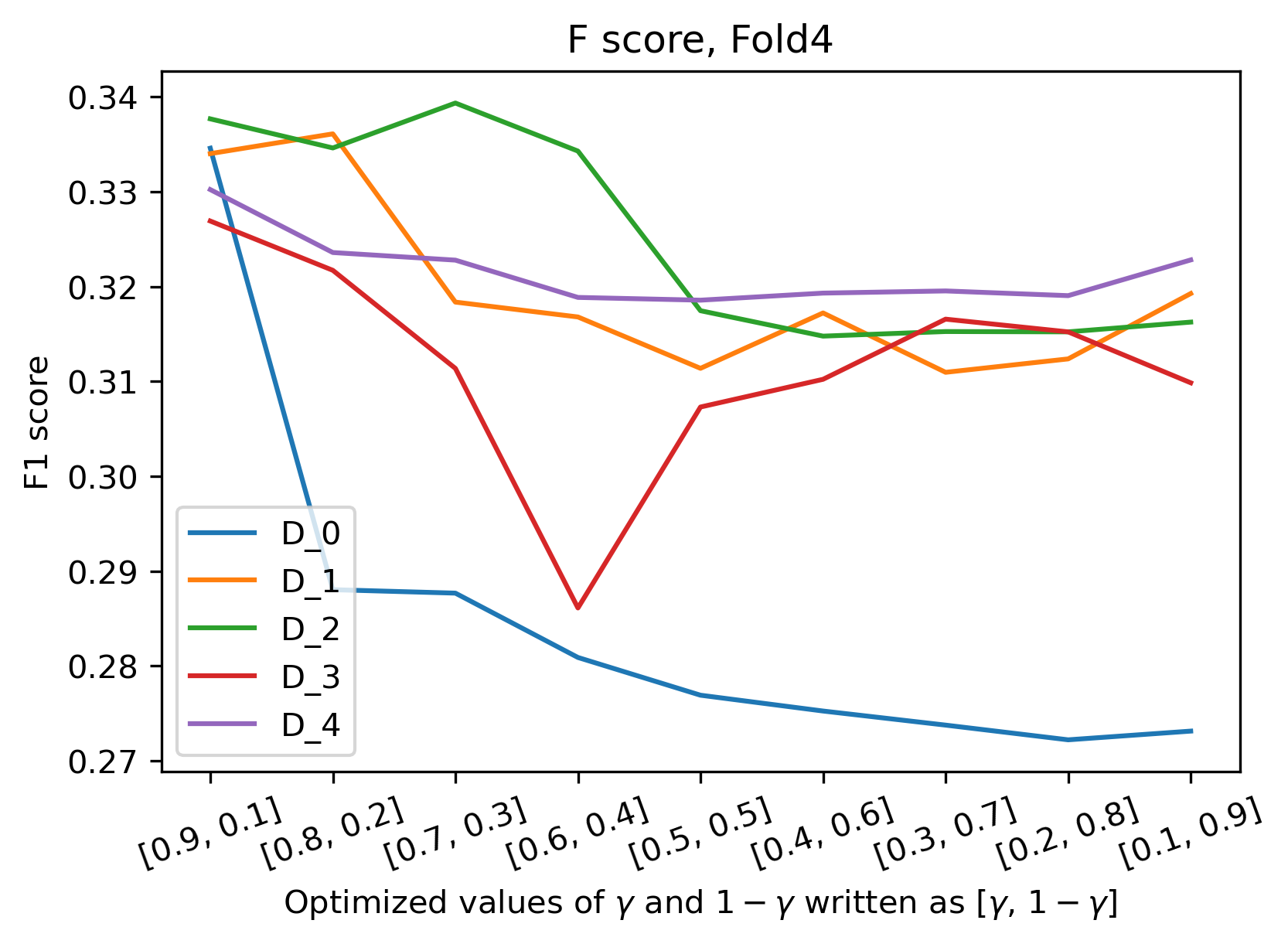}
    \caption{\tiny The changes in the F1 measure of validation set in Fold 4.}
  \end{subfigure}
  \caption{The performance of BSAC in the hyper-parameter selection phase of each Fold on the Lending Club data  as the parameter $\gamma$ decreases. The decline of classification performance based on F1 measure indicates that the the extra layer of Supervised Autoencoder plays a substantial role in obtaining a higher classification performance.}
  \label{gamma_folds_LC}
\end{figure}

Figure \ref{gamma_folds_LC} illustrates the performance of each base classifier (trained on balanced subsets of data $D_i$) on the validation set in each fold based on the F1 measure, which was used in the hyper-parameter selection. The decline of the performance on the validation set, as $\gamma$ is decreasing indicates that the Autoencoder's representations play a substantial role in the classification performance. 

\begin{figure}[ht]
  \centering
  \begin{subfigure}[b]{0.3\linewidth}
    \includegraphics[width=\linewidth]{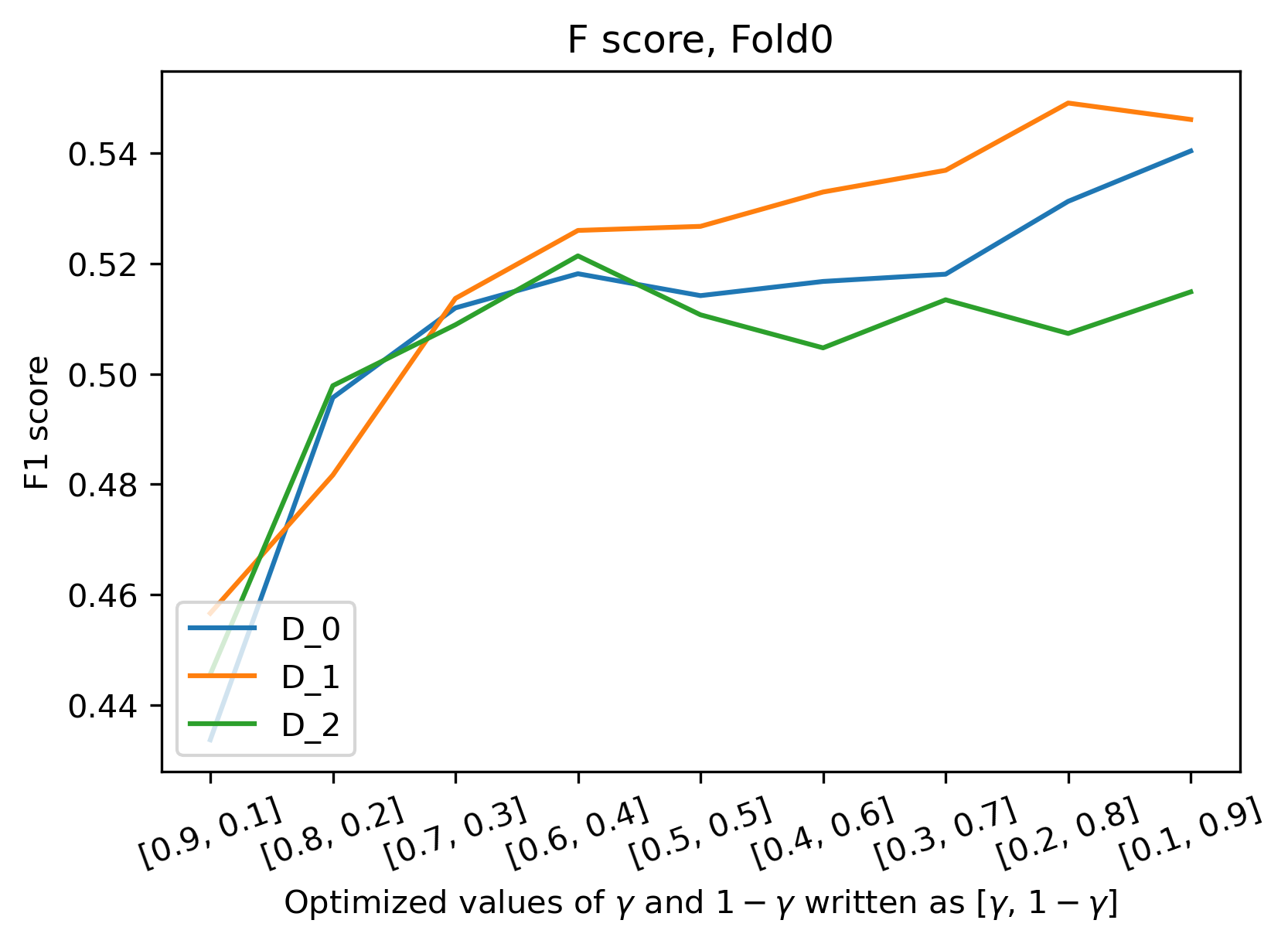}
    \caption{\tiny The changes in the F1 measure of validation set in Fold 0.}
  \end{subfigure}
  \begin{subfigure}[b]{0.3\linewidth}
    \includegraphics[width=\linewidth]{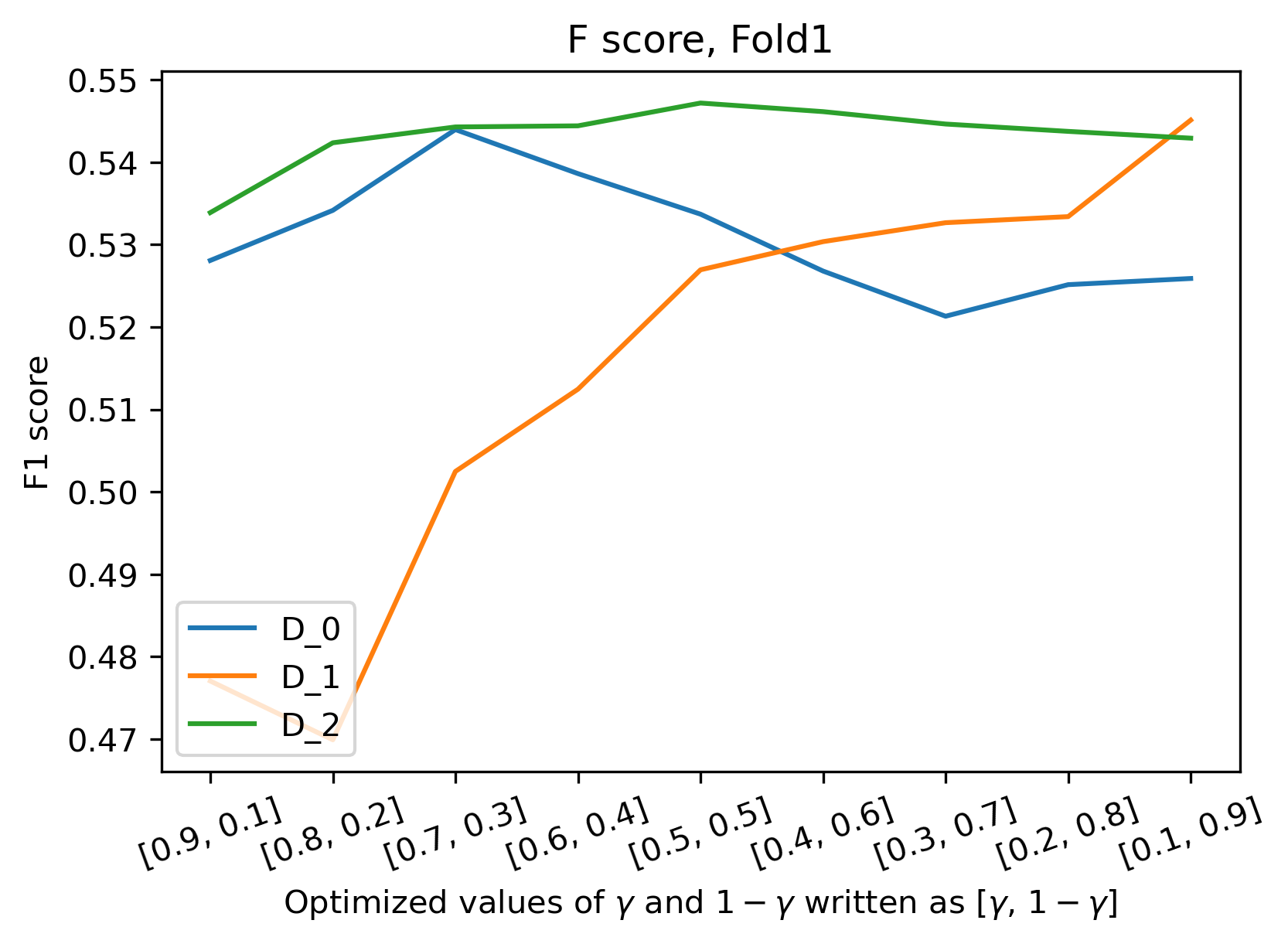}
    \caption{\tiny The changes in the F1 measure of validation set in Fold 1.}
  \end{subfigure}
  \begin{subfigure}[b]{0.3\linewidth}
    \includegraphics[width=\linewidth]{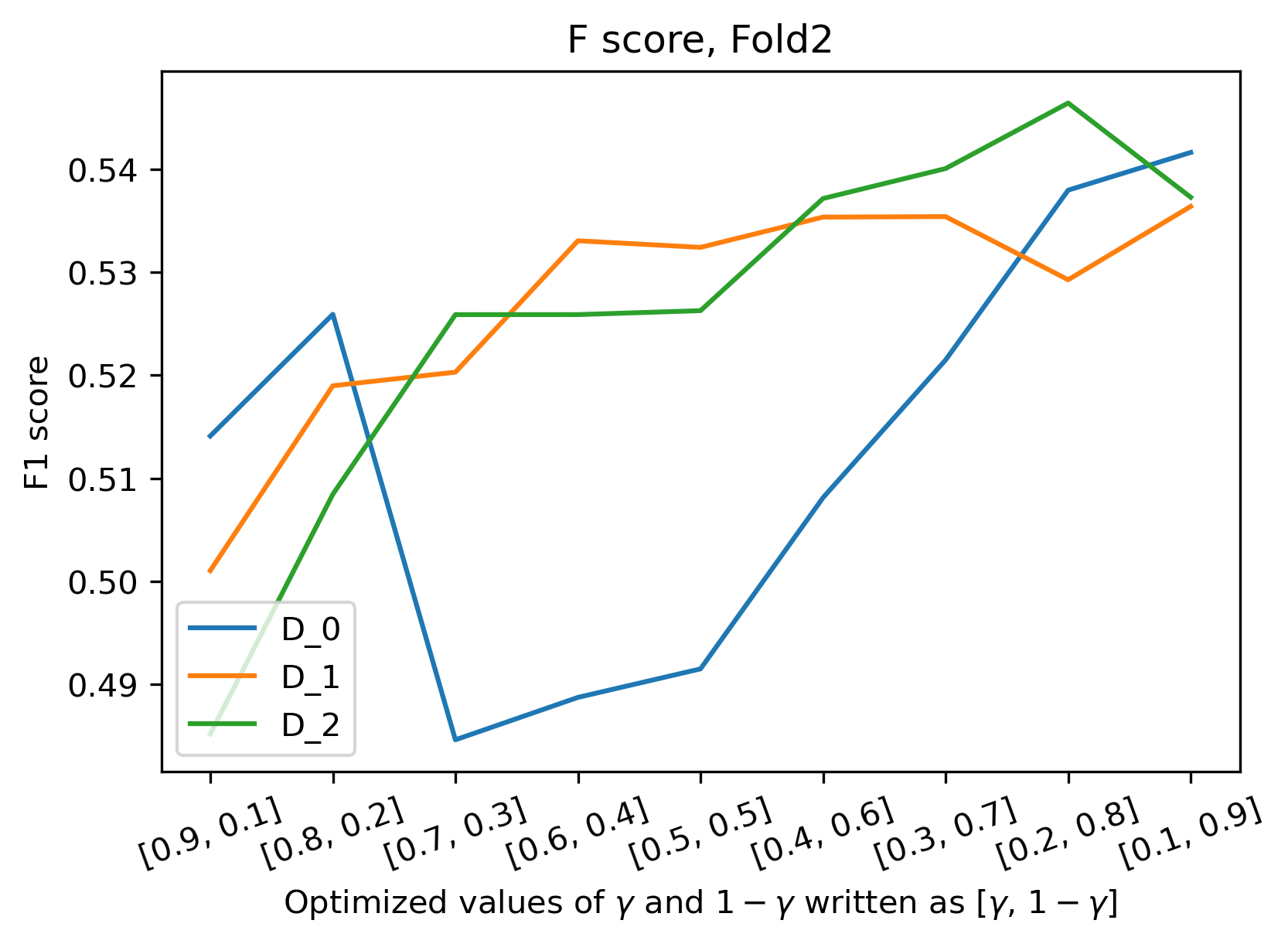}
    \caption{\tiny The changes in the F1 measure of validation set in Fold 2.}
  \end{subfigure}
  \begin{subfigure}[b]{0.3\linewidth}
    \includegraphics[width=\linewidth]{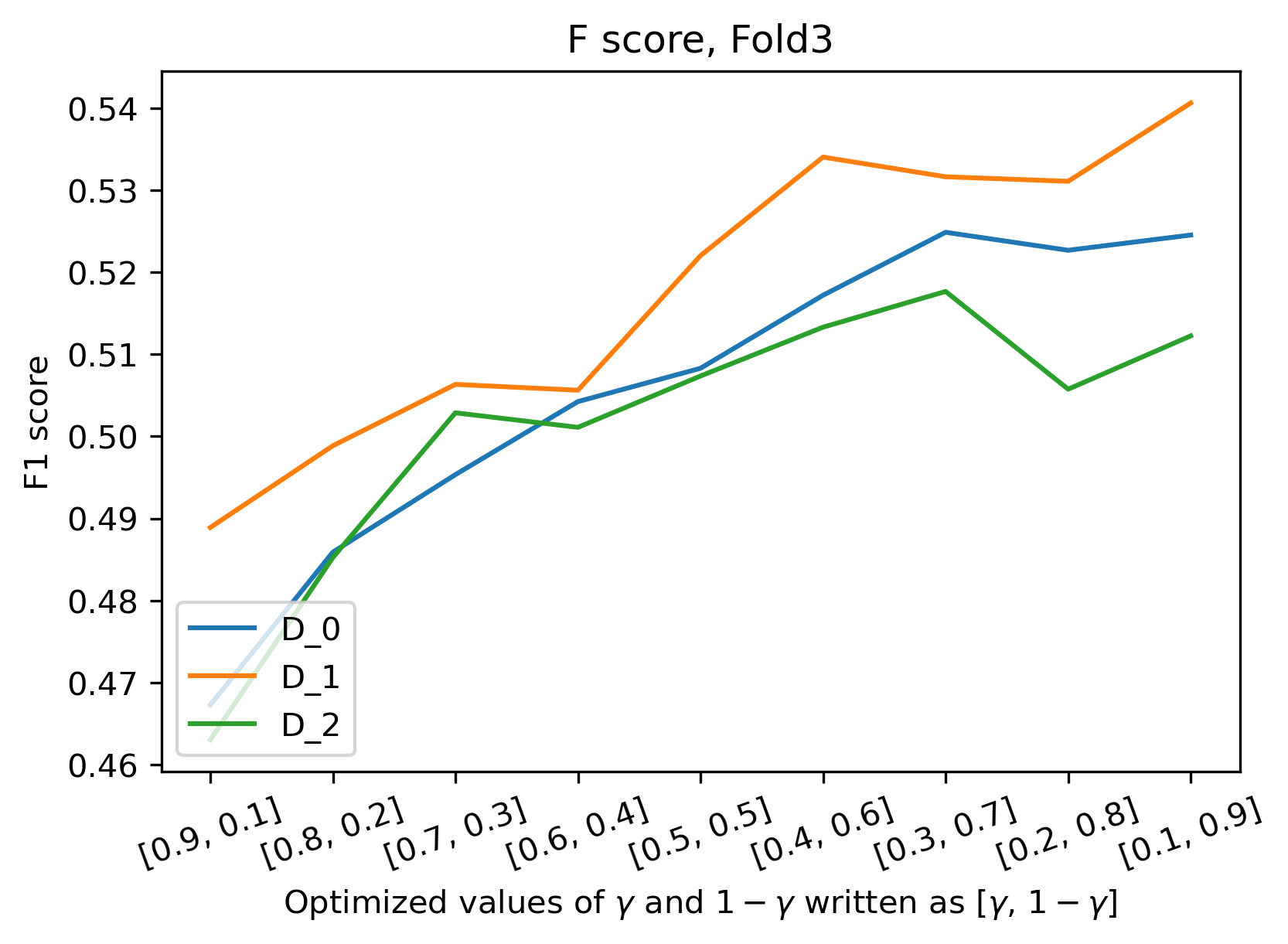}
    \caption{\tiny The changes in the F1 measure of validation set in Fold 3.}
  \end{subfigure}
  \begin{subfigure}[b]{0.3\linewidth}
    \includegraphics[width=\linewidth]{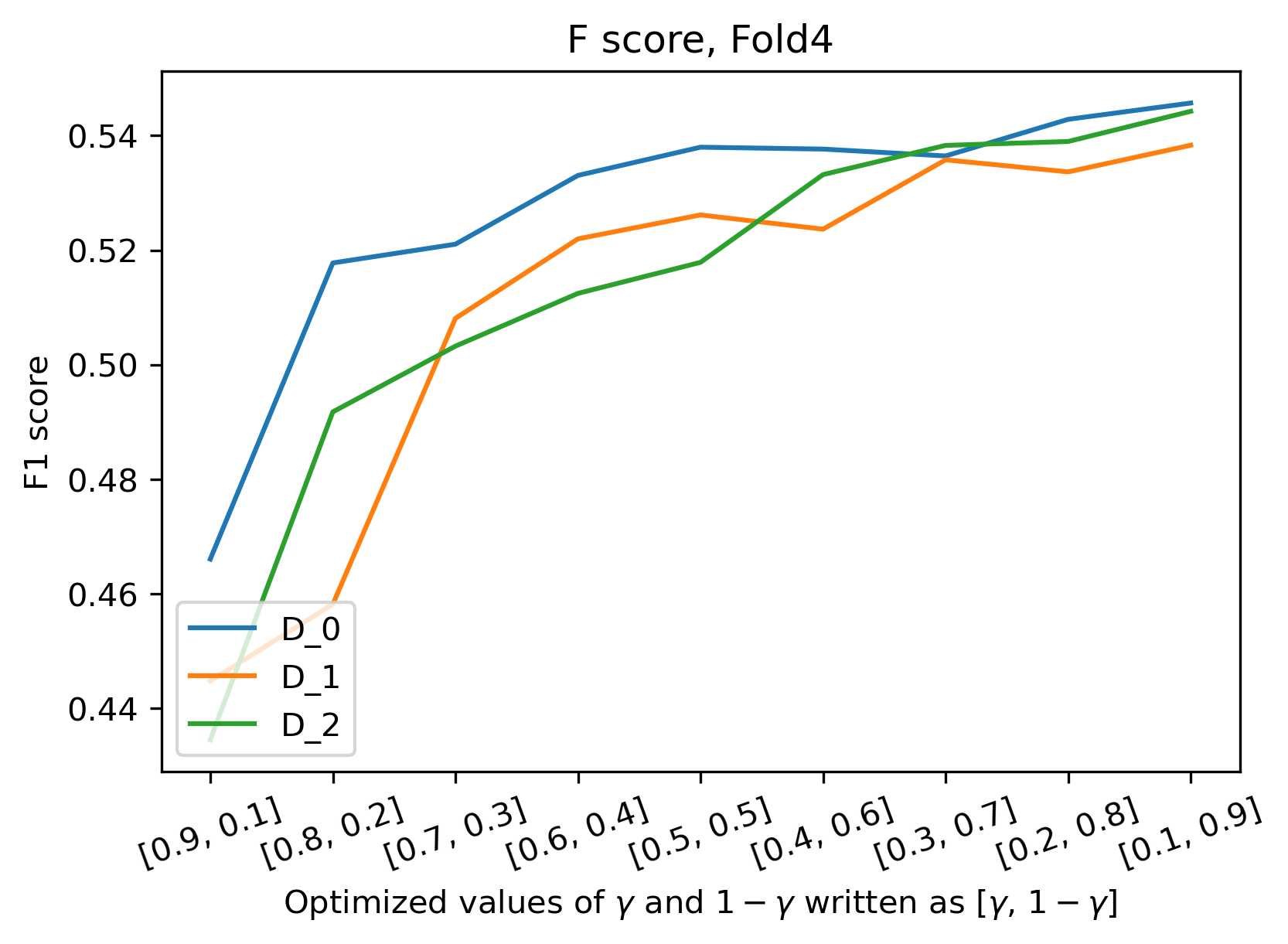}
    \caption{\tiny The changes in the F1 measure of validation set in Fold 4.}
  \end{subfigure}
  \caption{The performance of BSAC in the hyper-parameter selection phase of each Fold in cross validation on the benchmark data  as the parameter $\gamma$ decreases. The general decrease of classification performance based on F1 evaluation metric indicates that the input data reconstruction does not affect the classification performance substantially due to the lower complexity of the Taiwan dataset.}
  \label{gamma_folds_taiwan}
\end{figure}

Similarly, Figure \ref{gamma_folds_taiwan} illustrates the performance of BSAC with respect to the F1 measure on validation data on the Taiwan dataset. In contrast to the Lending Club dataset, the increasing slope of the plots indicates that the classification performance is not dependent on the high values of $\gamma$. This difference can be justified by different complexities of the Taiwan and Lending Club datasets. As mentioned before, the Lending Club dataset suffers from a high class imbalance, which can negatively affect the classification task. Consequently, a higher $\gamma$ value was selected in the validation process to extract useful patterns for classifying the unknown data by assigning a higher importance to the reconstruction loss $L_r$.

\newpage
\section{Conclusion}
Credit scoring can be considered as one of the most important measures that banks and financial institutes take to evaluate and decrease financial risk. In the credit scoring literature, it is well established that a minor improvement in the credit scoring models' performance can lead to significant savings in the financial sector. The class imbalance problem in the credit scoring datasets, as well as the heterogeneous nature of credit scoring data are among the challenges that limit the generalization power of credit scoring models to achieve higher performance. Therefore, in this study, we set out to develop a novel method in credit scoring classification by proposing a supervised Autoencoder network to embed the underlying information of credit scoring heterogeneous data into a lower-dimensional space and decrease the negative effects of class imbalance. Our proposed method, Bagging Supervised Autoencoder Classifier, combines the advantages of representation learning, ensemble learning, multi-task learning and under-sampling of negative samples, in order to develop a robust classification model for credit scoring, which inherently suffers from the class imbalance problem. The results of our experiments on benchmark data and a real-life credit scoring dataset proves the promising classification performance of BSAC.

Due to the our computational limitations in running the experiments, we have only focused on optimizing the hyper-parameter $\gamma$ in our model. It is clear that a more detailed hyper-parameter optimization would further improve the classification performance. 

Since the decision-making process of accepting or rejecting loan applications is highly prone to error and bias, in the future we intend to investigate the explainability and fairness of our proposed model in credit scoring to decrease bias and disparity towards customers, and increase transparency of the decision-making process of financial institutions.

\newpage
\appendix
\section{Lending Club Dataset variables}

\begin{longtable}{@{}p{110.0 pt}p{223.0 pt}@{}}
\caption{Variables used in this study.}
\label{LC_var}\\
\toprule
Variables                     & Description                                                                                                                                                                                              \\* \midrule
\endfirsthead
\multicolumn{2}{c}%
{{\bfseries Table \thetable\ continued from previous page}} \\
\toprule
Variables                     & Description                                                                                                                                                                                              \\* \midrule
\endhead
\bottomrule
\endfoot
\endlastfoot
loan\_amnt                  & Amount of the loan applied for by the borrower.                                                                                                                                               \\
acc\_now\_delinq                        & The number of accounts on which the borrower is now delinquent.                                                                                                                     \\
int\_rate                   & Interest Rate on the loan                                                                                                                                                                                \\
installment                 & The monthly payment owed by the borrower if the loan originates.                                                                                                                                         \\
annual\_inc                 & The self-reported annual income provided by the borrower during registration.                                                                                                                            \\
emp\_length                 & employment length in years.                                                                                                                                                                             \\
verification\_status        & Indicates if income was verified by LC~\footnote{Lending Club}, not verified, or if the income source was verified                                                                                                               \\
dti                         & The borrower’s total monthly debt payments on the total debt obligations, excluding mortgage and the requested LC loan, divided by the borrower’s self-reported monthly income. \\
delinq\_2yrs                & The number of 30+ days past-due incidences of delinquency in the borrower's credit file for the past 2 years                                                                                             \\
average\_fico            & The average value of the upper and lower boundary range of the borrower’s FICO at loan origination.                                                                                                                            \\
inq\_last\_6mths            & The number of inquiries in past 6 months (excluding auto and mortgage inquiries).                                                                                                                         \\
open\_acc                   & The number of open credit lines in the borrower's credit file.                                                                                                                                            \\
pub\_rec                    & Number of derogatory public records.                                                                                                                                                                      \\
revol\_util                 & Revolving line utilization rate, or the amount of credit the borrower is using relative to all available revolving credit.                                                                               \\
total\_acc                  & The total number of credit lines currently in the borrower's credit file.                                                                                                                                 \\
chargeoff\_within\_12\_mths & Number of charge-offs within 12 months.                                                                                                                                                                   \\
delinq\_amnt                & The past-due amount owed for the accounts on which the borrower is now delinquent                                                                                                                       \\
mort\_acc                   & Number of mortgage accounts                                                                                                                                                                             \\
pub\_rec\_bankruptcies      & Number of public record bankruptcies                                                                                                                                                                     \\
tax\_liens                  & Number of tax liens                                                                                                                                                                                     \\
grade                       & LC assigned loan grade. The values are: [A,B,C,D,E,F,G]                                                                                                                                                                                  \\
subgrade                    & LC assigned loan subgrade. The values are: [A1,\dots,A5,\dots,G1,\dots,G5]                                                                                                                                                                               \\
home\_ownership             & The home ownership status provided by the borrower during registration or obtained from the credit report. The values are: RENT, OWN, MORTGAGE                                                    \\
purpose                     & A category provided by the borrower for the loan request such as: Debt consolidation, small business, vacation, etc.                                                                                                                                                 \\
initial\_list\_status       & The initial listing status of the loan. The values are: Whole, Fractional                                                                                                                                       \\
loan\_status                & Current status of the loan (Target variable)                                                                                                                                                                               \\* \bottomrule
\end{longtable}

\bibliographystyle{apacite}
\bibliography{aereferences}

\begin{thebibliography}{}

\bibitem [\protect \citeauthoryear {%
Baesens%
\ \protect \BOthers {.}}{%
Baesens%
\ \protect \BOthers {.}}{%
{\protect \APACyear {2003}}%
}]{%
baesens2003benchmarking}
\APACinsertmetastar {%
baesens2003benchmarking}%
\begin{APACrefauthors}%
Baesens, B.%
, Van~Gestel, T.%
, Viaene, S.%
, Stepanova, M.%
, Suykens, J.%
\BCBL {}\ \BBA {} Vanthienen, J.%
\end{APACrefauthors}%
\unskip\
\newblock
\APACrefYearMonthDay{2003}{}{}.
\newblock
{\BBOQ}\APACrefatitle {Benchmarking state-of-the-art classification algorithms
  for credit scoring} {Benchmarking state-of-the-art classification algorithms
  for credit scoring}.{\BBCQ}
\newblock
\APACjournalVolNumPages{Journal of the operational research
  society}{54}{6}{627--635}.
\PrintBackRefs{\CurrentBib}

\bibitem [\protect \citeauthoryear {%
Bahnsen%
, Aouada%
\BCBL {}\ \BBA {} Ottersten%
}{%
Bahnsen%
\ \protect \BOthers {.}}{%
{\protect \APACyear {2015}}%
}]{%
bahnsen2015example}
\APACinsertmetastar {%
bahnsen2015example}%
\begin{APACrefauthors}%
Bahnsen, A\BPBI C.%
, Aouada, D.%
\BCBL {}\ \BBA {} Ottersten, B.%
\end{APACrefauthors}%
\unskip\
\newblock
\APACrefYearMonthDay{2015}{}{}.
\newblock
{\BBOQ}\APACrefatitle {Example-dependent cost-sensitive decision trees}
  {Example-dependent cost-sensitive decision trees}.{\BBCQ}
\newblock
\APACjournalVolNumPages{Expert Systems with Applications}{42}{19}{6609--6619}.
\PrintBackRefs{\CurrentBib}

\bibitem [\protect \citeauthoryear {%
Bastani%
, Asgari%
\BCBL {}\ \BBA {} Namavari%
}{%
Bastani%
\ \protect \BOthers {.}}{%
{\protect \APACyear {2019}}%
}]{%
bastani2019wide}
\APACinsertmetastar {%
bastani2019wide}%
\begin{APACrefauthors}%
Bastani, K.%
, Asgari, E.%
\BCBL {}\ \BBA {} Namavari, H.%
\end{APACrefauthors}%
\unskip\
\newblock
\APACrefYearMonthDay{2019}{}{}.
\newblock
{\BBOQ}\APACrefatitle {Wide and deep learning for peer-to-peer lending} {Wide
  and deep learning for peer-to-peer lending}.{\BBCQ}
\newblock
\APACjournalVolNumPages{Expert Systems with Applications}{134}{}{209--224}.
\PrintBackRefs{\CurrentBib}

\bibitem [\protect \citeauthoryear {%
Batuwita%
\ \BBA {} Palade%
}{%
Batuwita%
\ \BBA {} Palade%
}{%
{\protect \APACyear {2010}}%
}]{%
batuwita2010fsvm}
\APACinsertmetastar {%
batuwita2010fsvm}%
\begin{APACrefauthors}%
Batuwita, R.%
\BCBT {}\ \BBA {} Palade, V.%
\end{APACrefauthors}%
\unskip\
\newblock
\APACrefYearMonthDay{2010}{}{}.
\newblock
{\BBOQ}\APACrefatitle {FSVM-CIL: fuzzy support vector machines for class
  imbalance learning} {Fsvm-cil: fuzzy support vector machines for class
  imbalance learning}.{\BBCQ}
\newblock
\APACjournalVolNumPages{IEEE Transactions on Fuzzy Systems}{18}{3}{558--571}.
\PrintBackRefs{\CurrentBib}

\bibitem [\protect \citeauthoryear {%
Bengio%
, Courville%
\BCBL {}\ \BBA {} Vincent%
}{%
Bengio%
\ \protect \BOthers {.}}{%
{\protect \APACyear {2012}}%
}]{%
bengio2012unsupervised}
\APACinsertmetastar {%
bengio2012unsupervised}%
\begin{APACrefauthors}%
Bengio, Y.%
, Courville, A\BPBI C.%
\BCBL {}\ \BBA {} Vincent, P.%
\end{APACrefauthors}%
\unskip\
\newblock
\APACrefYearMonthDay{2012}{}{}.
\newblock
{\BBOQ}\APACrefatitle {Unsupervised feature learning and deep learning: A
  review and new perspectives} {Unsupervised feature learning and deep
  learning: A review and new perspectives}.{\BBCQ}
\newblock
\APACjournalVolNumPages{CoRR, abs/1206.5538}{1}{}{2012}.
\PrintBackRefs{\CurrentBib}

\bibitem [\protect \citeauthoryear {%
Bhatore%
, Mohan%
\BCBL {}\ \BBA {} Reddy%
}{%
Bhatore%
\ \protect \BOthers {.}}{%
{\protect \APACyear {2020}}%
}]{%
bhatore2020machine}
\APACinsertmetastar {%
bhatore2020machine}%
\begin{APACrefauthors}%
Bhatore, S.%
, Mohan, L.%
\BCBL {}\ \BBA {} Reddy, Y\BPBI R.%
\end{APACrefauthors}%
\unskip\
\newblock
\APACrefYearMonthDay{2020}{}{}.
\newblock
{\BBOQ}\APACrefatitle {Machine learning techniques for credit risk evaluation:
  a systematic literature review} {Machine learning techniques for credit risk
  evaluation: a systematic literature review}.{\BBCQ}
\newblock
\APACjournalVolNumPages{Journal of Banking and Financial
  Technology}{}{}{1--28}.
\PrintBackRefs{\CurrentBib}

\bibitem [\protect \citeauthoryear {%
Breiman%
}{%
Breiman%
}{%
{\protect \APACyear {1996}}%
}]{%
breiman1996bagging}
\APACinsertmetastar {%
breiman1996bagging}%
\begin{APACrefauthors}%
Breiman, L.%
\end{APACrefauthors}%
\unskip\
\newblock
\APACrefYearMonthDay{1996}{}{}.
\newblock
{\BBOQ}\APACrefatitle {Bagging predictors} {Bagging predictors}.{\BBCQ}
\newblock
\APACjournalVolNumPages{Machine learning}{24}{2}{123--140}.
\PrintBackRefs{\CurrentBib}

\bibitem [\protect \citeauthoryear {%
Brown%
\ \BBA {} Mues%
}{%
Brown%
\ \BBA {} Mues%
}{%
{\protect \APACyear {2012}}%
}]{%
brown2012experimental}
\APACinsertmetastar {%
brown2012experimental}%
\begin{APACrefauthors}%
Brown, I.%
\BCBT {}\ \BBA {} Mues, C.%
\end{APACrefauthors}%
\unskip\
\newblock
\APACrefYearMonthDay{2012}{}{}.
\newblock
{\BBOQ}\APACrefatitle {An experimental comparison of classification algorithms
  for imbalanced credit scoring data sets} {An experimental comparison of
  classification algorithms for imbalanced credit scoring data sets}.{\BBCQ}
\newblock
\APACjournalVolNumPages{Expert Systems with Applications}{39}{3}{3446--3453}.
\PrintBackRefs{\CurrentBib}

\bibitem [\protect \citeauthoryear {%
Caruana%
}{%
Caruana%
}{%
{\protect \APACyear {1997}}%
}]{%
caruana1997multitask}
\APACinsertmetastar {%
caruana1997multitask}%
\begin{APACrefauthors}%
Caruana, R.%
\end{APACrefauthors}%
\unskip\
\newblock
\APACrefYearMonthDay{1997}{}{}.
\newblock
{\BBOQ}\APACrefatitle {Multitask learning} {Multitask learning}.{\BBCQ}
\newblock
\APACjournalVolNumPages{Machine learning}{28}{1}{41--75}.
\PrintBackRefs{\CurrentBib}

\bibitem [\protect \citeauthoryear {%
Chawla%
, Bowyer%
, Hall%
\BCBL {}\ \BBA {} Kegelmeyer%
}{%
Chawla%
\ \protect \BOthers {.}}{%
{\protect \APACyear {2002}}%
}]{%
chawla2002smote}
\APACinsertmetastar {%
chawla2002smote}%
\begin{APACrefauthors}%
Chawla, N\BPBI V.%
, Bowyer, K\BPBI W.%
, Hall, L\BPBI O.%
\BCBL {}\ \BBA {} Kegelmeyer, W\BPBI P.%
\end{APACrefauthors}%
\unskip\
\newblock
\APACrefYearMonthDay{2002}{}{}.
\newblock
{\BBOQ}\APACrefatitle {SMOTE: synthetic minority over-sampling technique}
  {Smote: synthetic minority over-sampling technique}.{\BBCQ}
\newblock
\APACjournalVolNumPages{Journal of artificial intelligence
  research}{16}{}{321--357}.
\PrintBackRefs{\CurrentBib}

\bibitem [\protect \citeauthoryear {%
Chen%
, Wang%
\BCBL {}\ \BBA {} Liu%
}{%
Chen%
\ \protect \BOthers {.}}{%
{\protect \APACyear {2019}}%
}]{%
chen2019credit}
\APACinsertmetastar {%
chen2019credit}%
\begin{APACrefauthors}%
Chen, S.%
, Wang, Q.%
\BCBL {}\ \BBA {} Liu, S.%
\end{APACrefauthors}%
\unskip\
\newblock
\APACrefYearMonthDay{2019}{}{}.
\newblock
{\BBOQ}\APACrefatitle {Credit Risk Prediction in Peer-to-Peer Lending with
  Ensemble Learning Framework} {Credit risk prediction in peer-to-peer lending
  with ensemble learning framework}.{\BBCQ}
\newblock
\BIn{} \APACrefbtitle {2019 Chinese Control And Decision Conference (CCDC)}
  {2019 chinese control and decision conference (ccdc)}\ (\BPGS\ 4373--4377).
\PrintBackRefs{\CurrentBib}

\bibitem [\protect \citeauthoryear {%
Dastile%
, Celik%
\BCBL {}\ \BBA {} Potsane%
}{%
Dastile%
\ \protect \BOthers {.}}{%
{\protect \APACyear {2020}}%
}]{%
dastile2020statistical}
\APACinsertmetastar {%
dastile2020statistical}%
\begin{APACrefauthors}%
Dastile, X.%
, Celik, T.%
\BCBL {}\ \BBA {} Potsane, M.%
\end{APACrefauthors}%
\unskip\
\newblock
\APACrefYearMonthDay{2020}{}{}.
\newblock
{\BBOQ}\APACrefatitle {Statistical and machine learning models in credit
  scoring: A systematic literature survey} {Statistical and machine learning
  models in credit scoring: A systematic literature survey}.{\BBCQ}
\newblock
\APACjournalVolNumPages{Applied Soft Computing}{}{}{106263}.
\PrintBackRefs{\CurrentBib}

\bibitem [\protect \citeauthoryear {%
Duan%
}{%
Duan%
}{%
{\protect \APACyear {2019}}%
}]{%
duan2019financial}
\APACinsertmetastar {%
duan2019financial}%
\begin{APACrefauthors}%
Duan, J.%
\end{APACrefauthors}%
\unskip\
\newblock
\APACrefYearMonthDay{2019}{}{}.
\newblock
{\BBOQ}\APACrefatitle {Financial system modeling using deep neural networks
  (DNNs) for effective risk assessment and prediction} {Financial system
  modeling using deep neural networks (dnns) for effective risk assessment and
  prediction}.{\BBCQ}
\newblock
\APACjournalVolNumPages{Journal of the Franklin Institute}{356}{8}{4716--4731}.
\PrintBackRefs{\CurrentBib}

\bibitem [\protect \citeauthoryear {%
Duin%
}{%
Duin%
}{%
{\protect \APACyear {2002}}%
}]{%
duin2002combining}
\APACinsertmetastar {%
duin2002combining}%
\begin{APACrefauthors}%
Duin, R\BPBI P.%
\end{APACrefauthors}%
\unskip\
\newblock
\APACrefYearMonthDay{2002}{}{}.
\newblock
{\BBOQ}\APACrefatitle {The combining classifier: to train or not to train?}
  {The combining classifier: to train or not to train?}{\BBCQ}
\newblock
\BIn{} \APACrefbtitle {Object recognition supported by user interaction for
  service robots} {Object recognition supported by user interaction for service
  robots}\ (\BVOL~2, \BPGS\ 765--770).
\PrintBackRefs{\CurrentBib}

\bibitem [\protect \citeauthoryear {%
Emekter%
, Tu%
, Jirasakuldech%
\BCBL {}\ \BBA {} Lu%
}{%
Emekter%
\ \protect \BOthers {.}}{%
{\protect \APACyear {2015}}%
}]{%
emekter2015evaluating}
\APACinsertmetastar {%
emekter2015evaluating}%
\begin{APACrefauthors}%
Emekter, R.%
, Tu, Y.%
, Jirasakuldech, B.%
\BCBL {}\ \BBA {} Lu, M.%
\end{APACrefauthors}%
\unskip\
\newblock
\APACrefYearMonthDay{2015}{}{}.
\newblock
{\BBOQ}\APACrefatitle {Evaluating credit risk and loan performance in online
  Peer-to-Peer (P2P) lending} {Evaluating credit risk and loan performance in
  online peer-to-peer (p2p) lending}.{\BBCQ}
\newblock
\APACjournalVolNumPages{Applied Economics}{47}{1}{54--70}.
\PrintBackRefs{\CurrentBib}

\bibitem [\protect \citeauthoryear {%
Fan%
\ \BBA {} Yang%
}{%
Fan%
\ \BBA {} Yang%
}{%
{\protect \APACyear {2018}}%
}]{%
fan2018denoising}
\APACinsertmetastar {%
fan2018denoising}%
\begin{APACrefauthors}%
Fan, Q.%
\BCBT {}\ \BBA {} Yang, J.%
\end{APACrefauthors}%
\unskip\
\newblock
\APACrefYearMonthDay{2018}{}{}.
\newblock
{\BBOQ}\APACrefatitle {A denoising autoencoder approach for credit risk
  analysis} {A denoising autoencoder approach for credit risk analysis}.{\BBCQ}
\newblock
\BIn{} \APACrefbtitle {Proceedings of the 2018 International Conference on
  Computing and Artificial Intelligence} {Proceedings of the 2018 international
  conference on computing and artificial intelligence}\ (\BPGS\ 62--65).
\PrintBackRefs{\CurrentBib}

\bibitem [\protect \citeauthoryear {%
Feng%
, Xiao%
, Zhong%
, Qiu%
\BCBL {}\ \BBA {} Dong%
}{%
Feng%
\ \protect \BOthers {.}}{%
{\protect \APACyear {2018}}%
}]{%
feng2018dynamic}
\APACinsertmetastar {%
feng2018dynamic}%
\begin{APACrefauthors}%
Feng, X.%
, Xiao, Z.%
, Zhong, B.%
, Qiu, J.%
\BCBL {}\ \BBA {} Dong, Y.%
\end{APACrefauthors}%
\unskip\
\newblock
\APACrefYearMonthDay{2018}{}{}.
\newblock
{\BBOQ}\APACrefatitle {Dynamic ensemble classification for credit scoring using
  soft probability} {Dynamic ensemble classification for credit scoring using
  soft probability}.{\BBCQ}
\newblock
\APACjournalVolNumPages{Applied Soft Computing}{65}{}{139--151}.
\PrintBackRefs{\CurrentBib}

\bibitem [\protect \citeauthoryear {%
Galar%
, Fernandez%
, Barrenechea%
, Bustince%
\BCBL {}\ \BBA {} Herrera%
}{%
Galar%
\ \protect \BOthers {.}}{%
{\protect \APACyear {2011}}%
}]{%
galar2011review}
\APACinsertmetastar {%
galar2011review}%
\begin{APACrefauthors}%
Galar, M.%
, Fernandez, A.%
, Barrenechea, E.%
, Bustince, H.%
\BCBL {}\ \BBA {} Herrera, F.%
\end{APACrefauthors}%
\unskip\
\newblock
\APACrefYearMonthDay{2011}{}{}.
\newblock
{\BBOQ}\APACrefatitle {A review on ensembles for the class imbalance problem:
  bagging-, boosting-, and hybrid-based approaches} {A review on ensembles for
  the class imbalance problem: bagging-, boosting-, and hybrid-based
  approaches}.{\BBCQ}
\newblock
\APACjournalVolNumPages{IEEE Transactions on Systems, Man, and Cybernetics,
  Part C (Applications and Reviews)}{42}{4}{463--484}.
\PrintBackRefs{\CurrentBib}

\bibitem [\protect \citeauthoryear {%
Garc{\'\i}a%
, Marqu{\'e}s%
\BCBL {}\ \BBA {} S{\'a}nchez%
}{%
Garc{\'\i}a%
\ \protect \BOthers {.}}{%
{\protect \APACyear {2019}}%
}]{%
garcia2019exploring}
\APACinsertmetastar {%
garcia2019exploring}%
\begin{APACrefauthors}%
Garc{\'\i}a, V.%
, Marqu{\'e}s, A\BPBI I.%
\BCBL {}\ \BBA {} S{\'a}nchez, J\BPBI S.%
\end{APACrefauthors}%
\unskip\
\newblock
\APACrefYearMonthDay{2019}{}{}.
\newblock
{\BBOQ}\APACrefatitle {Exploring the synergetic effects of sample types on the
  performance of ensembles for credit risk and corporate bankruptcy prediction}
  {Exploring the synergetic effects of sample types on the performance of
  ensembles for credit risk and corporate bankruptcy prediction}.{\BBCQ}
\newblock
\APACjournalVolNumPages{Information Fusion}{47}{}{88--101}.
\PrintBackRefs{\CurrentBib}

\bibitem [\protect \citeauthoryear {%
Goodfellow%
, Bengio%
\BCBL {}\ \BBA {} Courville%
}{%
Goodfellow%
\ \protect \BOthers {.}}{%
{\protect \APACyear {2016}}%
}]{%
Goodfellow-et-al-2016}
\APACinsertmetastar {%
Goodfellow-et-al-2016}%
\begin{APACrefauthors}%
Goodfellow, I.%
, Bengio, Y.%
\BCBL {}\ \BBA {} Courville, A.%
\end{APACrefauthors}%
\unskip\
\newblock
\APACrefYear{2016}.
\newblock
\APACrefbtitle {Deep Learning} {Deep learning}.
\newblock
\APACaddressPublisher{}{MIT Press}.
\newblock
\APACrefnote{\url{http://www.deeplearningbook.org}}
\PrintBackRefs{\CurrentBib}

\bibitem [\protect \citeauthoryear {%
Haixiang%
\ \protect \BOthers {.}}{%
Haixiang%
\ \protect \BOthers {.}}{%
{\protect \APACyear {2017}}%
}]{%
haixiang2017learning}
\APACinsertmetastar {%
haixiang2017learning}%
\begin{APACrefauthors}%
Haixiang, G.%
, Yijing, L.%
, Shang, J.%
, Mingyun, G.%
, Yuanyue, H.%
\BCBL {}\ \BBA {} Bing, G.%
\end{APACrefauthors}%
\unskip\
\newblock
\APACrefYearMonthDay{2017}{}{}.
\newblock
{\BBOQ}\APACrefatitle {Learning from class-imbalanced data: Review of methods
  and applications} {Learning from class-imbalanced data: Review of methods and
  applications}.{\BBCQ}
\newblock
\APACjournalVolNumPages{Expert Systems with Applications}{73}{}{220--239}.
\PrintBackRefs{\CurrentBib}

\bibitem [\protect \citeauthoryear {%
He%
, Zhang%
\BCBL {}\ \BBA {} Zhang%
}{%
He%
\ \protect \BOthers {.}}{%
{\protect \APACyear {2018}}%
}]{%
he2018novel}
\APACinsertmetastar {%
he2018novel}%
\begin{APACrefauthors}%
He, H.%
, Zhang, W.%
\BCBL {}\ \BBA {} Zhang, S.%
\end{APACrefauthors}%
\unskip\
\newblock
\APACrefYearMonthDay{2018}{}{}.
\newblock
{\BBOQ}\APACrefatitle {A novel ensemble method for credit scoring: Adaption of
  different imbalance ratios} {A novel ensemble method for credit scoring:
  Adaption of different imbalance ratios}.{\BBCQ}
\newblock
\APACjournalVolNumPages{Expert Systems with Applications}{98}{}{105--117}.
\PrintBackRefs{\CurrentBib}

\bibitem [\protect \citeauthoryear {%
Kittler%
, Hatef%
, Duin%
\BCBL {}\ \BBA {} Matas%
}{%
Kittler%
\ \protect \BOthers {.}}{%
{\protect \APACyear {1998}}%
}]{%
kittler1998combining}
\APACinsertmetastar {%
kittler1998combining}%
\begin{APACrefauthors}%
Kittler, J.%
, Hatef, M.%
, Duin, R\BPBI P.%
\BCBL {}\ \BBA {} Matas, J.%
\end{APACrefauthors}%
\unskip\
\newblock
\APACrefYearMonthDay{1998}{}{}.
\newblock
{\BBOQ}\APACrefatitle {On combining classifiers} {On combining
  classifiers}.{\BBCQ}
\newblock
\APACjournalVolNumPages{IEEE transactions on pattern analysis and machine
  intelligence}{20}{3}{226--239}.
\PrintBackRefs{\CurrentBib}

\bibitem [\protect \citeauthoryear {%
Kozodoi%
, Lessmann%
, Papakonstantinou%
, Gatsoulis%
\BCBL {}\ \BBA {} Baesens%
}{%
Kozodoi%
\ \protect \BOthers {.}}{%
{\protect \APACyear {2019}}%
}]{%
kozodoi2019multi}
\APACinsertmetastar {%
kozodoi2019multi}%
\begin{APACrefauthors}%
Kozodoi, N.%
, Lessmann, S.%
, Papakonstantinou, K.%
, Gatsoulis, Y.%
\BCBL {}\ \BBA {} Baesens, B.%
\end{APACrefauthors}%
\unskip\
\newblock
\APACrefYearMonthDay{2019}{}{}.
\newblock
{\BBOQ}\APACrefatitle {A multi-objective approach for profit-driven feature
  selection in credit scoring} {A multi-objective approach for profit-driven
  feature selection in credit scoring}.{\BBCQ}
\newblock
\APACjournalVolNumPages{Decision support systems}{120}{}{106--117}.
\PrintBackRefs{\CurrentBib}

\bibitem [\protect \citeauthoryear {%
Le%
, Patterson%
\BCBL {}\ \BBA {} White%
}{%
Le%
\ \protect \BOthers {.}}{%
{\protect \APACyear {2018}}%
}]{%
le2018supervised}
\APACinsertmetastar {%
le2018supervised}%
\begin{APACrefauthors}%
Le, L.%
, Patterson, A.%
\BCBL {}\ \BBA {} White, M.%
\end{APACrefauthors}%
\unskip\
\newblock
\APACrefYearMonthDay{2018}{}{}.
\newblock
{\BBOQ}\APACrefatitle {Supervised autoencoders: Improving generalization
  performance with unsupervised regularizers} {Supervised autoencoders:
  Improving generalization performance with unsupervised regularizers}.{\BBCQ}
\newblock
\APACjournalVolNumPages{Advances in neural information processing
  systems}{31}{}{107--117}.
\PrintBackRefs{\CurrentBib}

\bibitem [\protect \citeauthoryear {%
Lei%
\ \protect \BOthers {.}}{%
Lei%
\ \protect \BOthers {.}}{%
{\protect \APACyear {2019}}%
}]{%
lei2019generative}
\APACinsertmetastar {%
lei2019generative}%
\begin{APACrefauthors}%
Lei, K.%
, Xie, Y.%
, Zhong, S.%
, Dai, J.%
, Yang, M.%
\BCBL {}\ \BBA {} Shen, Y.%
\end{APACrefauthors}%
\unskip\
\newblock
\APACrefYearMonthDay{2019}{}{}.
\newblock
{\BBOQ}\APACrefatitle {Generative adversarial fusion network for class
  imbalance credit scoring} {Generative adversarial fusion network for class
  imbalance credit scoring}.{\BBCQ}
\newblock
\APACjournalVolNumPages{Neural Computing and Applications}{}{}{1--12}.
\PrintBackRefs{\CurrentBib}

\bibitem [\protect \citeauthoryear {%
Lessmann%
, Baesens%
, Seow%
\BCBL {}\ \BBA {} Thomas%
}{%
Lessmann%
\ \protect \BOthers {.}}{%
{\protect \APACyear {2015}}%
}]{%
lessmann2015benchmarking}
\APACinsertmetastar {%
lessmann2015benchmarking}%
\begin{APACrefauthors}%
Lessmann, S.%
, Baesens, B.%
, Seow, H\BHBI V.%
\BCBL {}\ \BBA {} Thomas, L\BPBI C.%
\end{APACrefauthors}%
\unskip\
\newblock
\APACrefYearMonthDay{2015}{}{}.
\newblock
{\BBOQ}\APACrefatitle {Benchmarking state-of-the-art classification algorithms
  for credit scoring: An update of research} {Benchmarking state-of-the-art
  classification algorithms for credit scoring: An update of research}.{\BBCQ}
\newblock
\APACjournalVolNumPages{European Journal of Operational
  Research}{247}{1}{124--136}.
\PrintBackRefs{\CurrentBib}

\bibitem [\protect \citeauthoryear {%
Maalouf%
\ \BBA {} Trafalis%
}{%
Maalouf%
\ \BBA {} Trafalis%
}{%
{\protect \APACyear {2011}}%
}]{%
maalouf2011robust}
\APACinsertmetastar {%
maalouf2011robust}%
\begin{APACrefauthors}%
Maalouf, M.%
\BCBT {}\ \BBA {} Trafalis, T\BPBI B.%
\end{APACrefauthors}%
\unskip\
\newblock
\APACrefYearMonthDay{2011}{}{}.
\newblock
{\BBOQ}\APACrefatitle {Robust weighted kernel logistic regression in imbalanced
  and rare events data} {Robust weighted kernel logistic regression in
  imbalanced and rare events data}.{\BBCQ}
\newblock
\APACjournalVolNumPages{Computational Statistics \& Data
  Analysis}{55}{1}{168--183}.
\PrintBackRefs{\CurrentBib}

\bibitem [\protect \citeauthoryear {%
Maldonado%
, Pérez%
\BCBL {}\ \BBA {} Bravo%
}{%
Maldonado%
\ \protect \BOthers {.}}{%
{\protect \APACyear {2017}}%
}]{%
MALDONADO2017656}
\APACinsertmetastar {%
MALDONADO2017656}%
\begin{APACrefauthors}%
Maldonado, S.%
, Pérez, J.%
\BCBL {}\ \BBA {} Bravo, C.%
\end{APACrefauthors}%
\unskip\
\newblock
\APACrefYearMonthDay{2017}{}{}.
\newblock
{\BBOQ}\APACrefatitle {Cost-based feature selection for Support Vector
  Machines: An application in credit scoring} {Cost-based feature selection for
  support vector machines: An application in credit scoring}.{\BBCQ}
\newblock
\APACjournalVolNumPages{European Journal of Operational
  Research}{261}{2}{656-665}.
\newblock
\begin{APACrefURL}
  \url{https://www.sciencedirect.com/science/article/pii/S0377221717301595}
  \end{APACrefURL}
\newblock
\begin{APACrefDOI} \doi{https://doi.org/10.1016/j.ejor.2017.02.037}
  \end{APACrefDOI}
\PrintBackRefs{\CurrentBib}

\bibitem [\protect \citeauthoryear {%
Malekipirbazari%
\ \BBA {} Aksakalli%
}{%
Malekipirbazari%
\ \BBA {} Aksakalli%
}{%
{\protect \APACyear {2015}}%
}]{%
malekipirbazari2015risk}
\APACinsertmetastar {%
malekipirbazari2015risk}%
\begin{APACrefauthors}%
Malekipirbazari, M.%
\BCBT {}\ \BBA {} Aksakalli, V.%
\end{APACrefauthors}%
\unskip\
\newblock
\APACrefYearMonthDay{2015}{}{}.
\newblock
{\BBOQ}\APACrefatitle {Risk assessment in social lending via random forests}
  {Risk assessment in social lending via random forests}.{\BBCQ}
\newblock
\APACjournalVolNumPages{Expert Systems with Applications}{42}{10}{4621--4631}.
\PrintBackRefs{\CurrentBib}

\bibitem [\protect \citeauthoryear {%
Mancisidor%
, Kampffmeyer%
, Aas%
\BCBL {}\ \BBA {} Jenssen%
}{%
Mancisidor%
\ \protect \BOthers {.}}{%
{\protect \APACyear {2018}}%
}]{%
mancisidor2018segment}
\APACinsertmetastar {%
mancisidor2018segment}%
\begin{APACrefauthors}%
Mancisidor, R\BPBI A.%
, Kampffmeyer, M.%
, Aas, K.%
\BCBL {}\ \BBA {} Jenssen, R.%
\end{APACrefauthors}%
\unskip\
\newblock
\APACrefYearMonthDay{2018}{}{}.
\newblock
{\BBOQ}\APACrefatitle {Segment-Based Credit Scoring Using Latent Clusters in
  the Variational Autoencoder} {Segment-based credit scoring using latent
  clusters in the variational autoencoder}.{\BBCQ}
\newblock
\APACjournalVolNumPages{arXiv preprint arXiv:1806.02538}{}{}{}.
\PrintBackRefs{\CurrentBib}

\bibitem [\protect \citeauthoryear {%
Najafabadi%
\ \protect \BOthers {.}}{%
Najafabadi%
\ \protect \BOthers {.}}{%
{\protect \APACyear {2015}}%
}]{%
najafabadi2015deep}
\APACinsertmetastar {%
najafabadi2015deep}%
\begin{APACrefauthors}%
Najafabadi, M\BPBI M.%
, Villanustre, F.%
, Khoshgoftaar, T\BPBI M.%
, Seliya, N.%
, Wald, R.%
\BCBL {}\ \BBA {} Muharemagic, E.%
\end{APACrefauthors}%
\unskip\
\newblock
\APACrefYearMonthDay{2015}{}{}.
\newblock
{\BBOQ}\APACrefatitle {Deep learning applications and challenges in big data
  analytics} {Deep learning applications and challenges in big data
  analytics}.{\BBCQ}
\newblock
\APACjournalVolNumPages{Journal of Big Data}{2}{1}{1}.
\PrintBackRefs{\CurrentBib}

\bibitem [\protect \citeauthoryear {%
Neagoe%
, Ciotec%
\BCBL {}\ \BBA {} Cucu%
}{%
Neagoe%
\ \protect \BOthers {.}}{%
{\protect \APACyear {2018}}%
}]{%
neagoe2018deep}
\APACinsertmetastar {%
neagoe2018deep}%
\begin{APACrefauthors}%
Neagoe, V\BHBI E.%
, Ciotec, A\BHBI D.%
\BCBL {}\ \BBA {} Cucu, G\BHBI S.%
\end{APACrefauthors}%
\unskip\
\newblock
\APACrefYearMonthDay{2018}{}{}.
\newblock
{\BBOQ}\APACrefatitle {Deep convolutional neural networks versus multilayer
  perceptron for financial prediction} {Deep convolutional neural networks
  versus multilayer perceptron for financial prediction}.{\BBCQ}
\newblock
\BIn{} \APACrefbtitle {2018 International Conference on Communications (COMM)}
  {2018 international conference on communications (comm)}\ (\BPGS\ 201--206).
\PrintBackRefs{\CurrentBib}

\bibitem [\protect \citeauthoryear {%
Papouskova%
\ \BBA {} Hajek%
}{%
Papouskova%
\ \BBA {} Hajek%
}{%
{\protect \APACyear {2019}}%
}]{%
papouskova2019two}
\APACinsertmetastar {%
papouskova2019two}%
\begin{APACrefauthors}%
Papouskova, M.%
\BCBT {}\ \BBA {} Hajek, P.%
\end{APACrefauthors}%
\unskip\
\newblock
\APACrefYearMonthDay{2019}{}{}.
\newblock
{\BBOQ}\APACrefatitle {Two-stage consumer credit risk modelling using
  heterogeneous ensemble learning} {Two-stage consumer credit risk modelling
  using heterogeneous ensemble learning}.{\BBCQ}
\newblock
\APACjournalVolNumPages{Decision Support Systems}{118}{}{33--45}.
\PrintBackRefs{\CurrentBib}

\bibitem [\protect \citeauthoryear {%
Reichert%
, Cho%
\BCBL {}\ \BBA {} Wagner%
}{%
Reichert%
\ \protect \BOthers {.}}{%
{\protect \APACyear {1983}}%
}]{%
reichert1983examination}
\APACinsertmetastar {%
reichert1983examination}%
\begin{APACrefauthors}%
Reichert, A\BPBI K.%
, Cho, C\BHBI C.%
\BCBL {}\ \BBA {} Wagner, G\BPBI M.%
\end{APACrefauthors}%
\unskip\
\newblock
\APACrefYearMonthDay{1983}{}{}.
\newblock
{\BBOQ}\APACrefatitle {An examination of the conceptual issues involved in
  developing credit-scoring models} {An examination of the conceptual issues
  involved in developing credit-scoring models}.{\BBCQ}
\newblock
\APACjournalVolNumPages{Journal of Business \& Economic
  Statistics}{1}{2}{101--114}.
\PrintBackRefs{\CurrentBib}

\bibitem [\protect \citeauthoryear {%
Serrano-Cinca%
, Guti{\'e}rrez-Nieto%
\BCBL {}\ \BBA {} L{\'o}pez-Palacios%
}{%
Serrano-Cinca%
\ \protect \BOthers {.}}{%
{\protect \APACyear {2015}}%
}]{%
serrano2015determinants}
\APACinsertmetastar {%
serrano2015determinants}%
\begin{APACrefauthors}%
Serrano-Cinca, C.%
, Guti{\'e}rrez-Nieto, B.%
\BCBL {}\ \BBA {} L{\'o}pez-Palacios, L.%
\end{APACrefauthors}%
\unskip\
\newblock
\APACrefYearMonthDay{2015}{}{}.
\newblock
{\BBOQ}\APACrefatitle {Determinants of default in P2P lending} {Determinants of
  default in p2p lending}.{\BBCQ}
\newblock
\APACjournalVolNumPages{PloS one}{10}{10}{e0139427}.
\PrintBackRefs{\CurrentBib}

\bibitem [\protect \citeauthoryear {%
Shen%
, Zhao%
, Kou%
\BCBL {}\ \BBA {} Alsaadi%
}{%
Shen%
\ \protect \BOthers {.}}{%
{\protect \APACyear {2020}}%
}]{%
shen2020new}
\APACinsertmetastar {%
shen2020new}%
\begin{APACrefauthors}%
Shen, F.%
, Zhao, X.%
, Kou, G.%
\BCBL {}\ \BBA {} Alsaadi, F\BPBI E.%
\end{APACrefauthors}%
\unskip\
\newblock
\APACrefYearMonthDay{2020}{}{}.
\newblock
{\BBOQ}\APACrefatitle {A new deep learning ensemble credit risk evaluation
  model with an improved synthetic minority oversampling technique} {A new deep
  learning ensemble credit risk evaluation model with an improved synthetic
  minority oversampling technique}.{\BBCQ}
\newblock
\APACjournalVolNumPages{Applied Soft Computing}{}{}{106852}.
\PrintBackRefs{\CurrentBib}

\bibitem [\protect \citeauthoryear {%
Sun%
, Lang%
, Fujita%
\BCBL {}\ \BBA {} Li%
}{%
Sun%
\ \protect \BOthers {.}}{%
{\protect \APACyear {2018}}%
}]{%
sun2018imbalanced}
\APACinsertmetastar {%
sun2018imbalanced}%
\begin{APACrefauthors}%
Sun, J.%
, Lang, J.%
, Fujita, H.%
\BCBL {}\ \BBA {} Li, H.%
\end{APACrefauthors}%
\unskip\
\newblock
\APACrefYearMonthDay{2018}{}{}.
\newblock
{\BBOQ}\APACrefatitle {Imbalanced enterprise credit evaluation with DTE-SBD:
  Decision tree ensemble based on SMOTE and bagging with differentiated
  sampling rates} {Imbalanced enterprise credit evaluation with dte-sbd:
  Decision tree ensemble based on smote and bagging with differentiated
  sampling rates}.{\BBCQ}
\newblock
\APACjournalVolNumPages{Information Sciences}{425}{}{76--91}.
\PrintBackRefs{\CurrentBib}

\bibitem [\protect \citeauthoryear {%
Tran%
, Duong%
\BCBL {}\ \BBA {} Ho%
}{%
Tran%
\ \protect \BOthers {.}}{%
{\protect \APACyear {2016}}%
}]{%
tran2016credit}
\APACinsertmetastar {%
tran2016credit}%
\begin{APACrefauthors}%
Tran, K.%
, Duong, T.%
\BCBL {}\ \BBA {} Ho, Q.%
\end{APACrefauthors}%
\unskip\
\newblock
\APACrefYearMonthDay{2016}{}{}.
\newblock
{\BBOQ}\APACrefatitle {Credit scoring model: A combination of genetic
  programming and deep learning} {Credit scoring model: A combination of
  genetic programming and deep learning}.{\BBCQ}
\newblock
\BIn{} \APACrefbtitle {2016 Future Technologies Conference (FTC)} {2016 future
  technologies conference (ftc)}\ (\BPGS\ 145--149).
\PrintBackRefs{\CurrentBib}

\bibitem [\protect \citeauthoryear {%
Waibel%
, Sawai%
\BCBL {}\ \BBA {} Shikano%
}{%
Waibel%
\ \protect \BOthers {.}}{%
{\protect \APACyear {1989}}%
}]{%
waibel1989modularity}
\APACinsertmetastar {%
waibel1989modularity}%
\begin{APACrefauthors}%
Waibel, A.%
, Sawai, H.%
\BCBL {}\ \BBA {} Shikano, K.%
\end{APACrefauthors}%
\unskip\
\newblock
\APACrefYearMonthDay{1989}{}{}.
\newblock
{\BBOQ}\APACrefatitle {Modularity and scaling in large phonemic neural
  networks} {Modularity and scaling in large phonemic neural networks}.{\BBCQ}
\newblock
\APACjournalVolNumPages{IEEE Transactions on Acoustics, Speech, and Signal
  Processing}{37}{12}{1888--1898}.
\PrintBackRefs{\CurrentBib}

\bibitem [\protect \citeauthoryear {%
Wang%
, Hao%
, Ma%
\BCBL {}\ \BBA {} Jiang%
}{%
Wang%
\ \protect \BOthers {.}}{%
{\protect \APACyear {2011}}%
}]{%
wang2011comparative}
\APACinsertmetastar {%
wang2011comparative}%
\begin{APACrefauthors}%
Wang, G.%
, Hao, J.%
, Ma, J.%
\BCBL {}\ \BBA {} Jiang, H.%
\end{APACrefauthors}%
\unskip\
\newblock
\APACrefYearMonthDay{2011}{}{}.
\newblock
{\BBOQ}\APACrefatitle {A comparative assessment of ensemble learning for credit
  scoring} {A comparative assessment of ensemble learning for credit
  scoring}.{\BBCQ}
\newblock
\APACjournalVolNumPages{Expert systems with applications}{38}{1}{223--230}.
\PrintBackRefs{\CurrentBib}

\bibitem [\protect \citeauthoryear {%
Wong%
, Seng%
\BCBL {}\ \BBA {} Wong%
}{%
Wong%
\ \protect \BOthers {.}}{%
{\protect \APACyear {2020}}%
}]{%
wong2020cost}
\APACinsertmetastar {%
wong2020cost}%
\begin{APACrefauthors}%
Wong, M\BPBI L.%
, Seng, K.%
\BCBL {}\ \BBA {} Wong, P\BPBI K.%
\end{APACrefauthors}%
\unskip\
\newblock
\APACrefYearMonthDay{2020}{}{}.
\newblock
{\BBOQ}\APACrefatitle {Cost-sensitive ensemble of stacked denoising
  autoencoders for class imbalance problems in business domain} {Cost-sensitive
  ensemble of stacked denoising autoencoders for class imbalance problems in
  business domain}.{\BBCQ}
\newblock
\APACjournalVolNumPages{Expert Systems with Applications}{141}{}{112918}.
\PrintBackRefs{\CurrentBib}

\bibitem [\protect \citeauthoryear {%
Xia%
, Liu%
, Da%
\BCBL {}\ \BBA {} Xie%
}{%
Xia%
\ \protect \BOthers {.}}{%
{\protect \APACyear {2018}}%
}]{%
xia2018novel}
\APACinsertmetastar {%
xia2018novel}%
\begin{APACrefauthors}%
Xia, Y.%
, Liu, C.%
, Da, B.%
\BCBL {}\ \BBA {} Xie, F.%
\end{APACrefauthors}%
\unskip\
\newblock
\APACrefYearMonthDay{2018}{}{}.
\newblock
{\BBOQ}\APACrefatitle {A novel heterogeneous ensemble credit scoring model
  based on bstacking approach} {A novel heterogeneous ensemble credit scoring
  model based on bstacking approach}.{\BBCQ}
\newblock
\APACjournalVolNumPages{Expert Systems with Applications}{93}{}{182--199}.
\PrintBackRefs{\CurrentBib}

\bibitem [\protect \citeauthoryear {%
Xia%
, Liu%
\BCBL {}\ \BBA {} Liu%
}{%
Xia%
\ \protect \BOthers {.}}{%
{\protect \APACyear {2017}}%
}]{%
xia2017cost}
\APACinsertmetastar {%
xia2017cost}%
\begin{APACrefauthors}%
Xia, Y.%
, Liu, C.%
\BCBL {}\ \BBA {} Liu, N.%
\end{APACrefauthors}%
\unskip\
\newblock
\APACrefYearMonthDay{2017}{}{}.
\newblock
{\BBOQ}\APACrefatitle {Cost-sensitive boosted tree for loan evaluation in
  peer-to-peer lending} {Cost-sensitive boosted tree for loan evaluation in
  peer-to-peer lending}.{\BBCQ}
\newblock
\APACjournalVolNumPages{Electronic Commerce Research and
  Applications}{24}{}{30--49}.
\PrintBackRefs{\CurrentBib}

\bibitem [\protect \citeauthoryear {%
H.~Xiao%
, Xiao%
\BCBL {}\ \BBA {} Wang%
}{%
H.~Xiao%
\ \protect \BOthers {.}}{%
{\protect \APACyear {2016}}%
}]{%
xiao2016ensemble}
\APACinsertmetastar {%
xiao2016ensemble}%
\begin{APACrefauthors}%
Xiao, H.%
, Xiao, Z.%
\BCBL {}\ \BBA {} Wang, Y.%
\end{APACrefauthors}%
\unskip\
\newblock
\APACrefYearMonthDay{2016}{}{}.
\newblock
{\BBOQ}\APACrefatitle {Ensemble classification based on supervised clustering
  for credit scoring} {Ensemble classification based on supervised clustering
  for credit scoring}.{\BBCQ}
\newblock
\APACjournalVolNumPages{Applied Soft Computing}{43}{}{73--86}.
\PrintBackRefs{\CurrentBib}

\bibitem [\protect \citeauthoryear {%
J.~Xiao%
, Xie%
, He%
\BCBL {}\ \BBA {} Jiang%
}{%
J.~Xiao%
\ \protect \BOthers {.}}{%
{\protect \APACyear {2012}}%
}]{%
xiao2012dynamic}
\APACinsertmetastar {%
xiao2012dynamic}%
\begin{APACrefauthors}%
Xiao, J.%
, Xie, L.%
, He, C.%
\BCBL {}\ \BBA {} Jiang, X.%
\end{APACrefauthors}%
\unskip\
\newblock
\APACrefYearMonthDay{2012}{}{}.
\newblock
{\BBOQ}\APACrefatitle {Dynamic classifier ensemble model for customer
  classification with imbalanced class distribution} {Dynamic classifier
  ensemble model for customer classification with imbalanced class
  distribution}.{\BBCQ}
\newblock
\APACjournalVolNumPages{Expert Systems with Applications}{39}{3}{3668--3675}.
\PrintBackRefs{\CurrentBib}

\bibitem [\protect \citeauthoryear {%
J.~Xiao%
\ \protect \BOthers {.}}{%
J.~Xiao%
\ \protect \BOthers {.}}{%
{\protect \APACyear {2020}}%
}]{%
xiao2020cost}
\APACinsertmetastar {%
xiao2020cost}%
\begin{APACrefauthors}%
Xiao, J.%
, Zhou, X.%
, Zhong, Y.%
, Xie, L.%
, Gu, X.%
\BCBL {}\ \BBA {} Liu, D.%
\end{APACrefauthors}%
\unskip\
\newblock
\APACrefYearMonthDay{2020}{}{}.
\newblock
{\BBOQ}\APACrefatitle {Cost-sensitive semi-supervised selective ensemble model
  for customer credit scoring} {Cost-sensitive semi-supervised selective
  ensemble model for customer credit scoring}.{\BBCQ}
\newblock
\APACjournalVolNumPages{Knowledge-Based Systems}{189}{}{105118}.
\PrintBackRefs{\CurrentBib}

\bibitem [\protect \citeauthoryear {%
Yu%
, Zhou%
, Tang%
\BCBL {}\ \BBA {} Chen%
}{%
Yu%
\ \protect \BOthers {.}}{%
{\protect \APACyear {2018}}%
}]{%
yu2018dbn}
\APACinsertmetastar {%
yu2018dbn}%
\begin{APACrefauthors}%
Yu, L.%
, Zhou, R.%
, Tang, L.%
\BCBL {}\ \BBA {} Chen, R.%
\end{APACrefauthors}%
\unskip\
\newblock
\APACrefYearMonthDay{2018}{}{}.
\newblock
{\BBOQ}\APACrefatitle {A DBN-based resampling SVM ensemble learning paradigm
  for credit classification with imbalanced data} {A dbn-based resampling svm
  ensemble learning paradigm for credit classification with imbalanced
  data}.{\BBCQ}
\newblock
\APACjournalVolNumPages{Applied Soft Computing}{69}{}{192--202}.
\PrintBackRefs{\CurrentBib}

\bibitem [\protect \citeauthoryear {%
Zhang%
\ \BBA {} Yang%
}{%
Zhang%
\ \BBA {} Yang%
}{%
{\protect \APACyear {2018}}%
}]{%
zhang2018overview}
\APACinsertmetastar {%
zhang2018overview}%
\begin{APACrefauthors}%
Zhang, Y.%
\BCBT {}\ \BBA {} Yang, Q.%
\end{APACrefauthors}%
\unskip\
\newblock
\APACrefYearMonthDay{2018}{}{}.
\newblock
{\BBOQ}\APACrefatitle {An overview of multi-task learning} {An overview of
  multi-task learning}.{\BBCQ}
\newblock
\APACjournalVolNumPages{National Science Review}{5}{1}{30--43}.
\PrintBackRefs{\CurrentBib}

\end{thebibliography}
\end{document}